\documentclass[12pt]{article}
\usepackage[T1]{fontenc}
\usepackage{lmodern}

\usepackage{PRIMEarxiv}

\usepackage{microtype}
\usepackage{graphicx}
\usepackage{subfigure}
\usepackage{booktabs} 

\usepackage{hyperref}

\usepackage[utf8]{inputenc}

\usepackage{amsmath}
\usepackage{amssymb}
\usepackage{mathtools}
\usepackage{amsthm}
\usepackage{dsfont}

\usepackage[capitalize,noabbrev]{cleveref}

\theoremstyle{plain}
\newtheorem{theorem}{Theorem}[section]

\newtheorem{lemma}[theorem]{Lemma}

\theoremstyle{definition}
\newtheorem{definition}[theorem]{Definition}
\newtheorem{assumption}[theorem]{Assumption}
\theoremstyle{remark}
\newtheorem{remark}[theorem]{Remark}

\usepackage{microtype}
\usepackage{graphicx}
\usepackage{subfigure}
\usepackage{booktabs} 
\usepackage{booktabs, multirow} 
\usepackage{soul}
\usepackage[table]{xcolor} 
\usepackage{changepage,threeparttable} 

\usepackage{booktabs, multirow} 
\usepackage{soul}
\usepackage[table]{xcolor} 

\usepackage{amsmath}

\DeclareMathOperator{\E}{\mathbb{E}}
\DeclareMathOperator{\KL}{\text{KL}}
\DeclareMathOperator{\ER}{\text{ER}}
\DeclareMathOperator{\CATE}{\text{CATE}}

\newcommand\independent{\perp \!\!\! \perp }

\usepackage{amssymb}

\usepackage[utf8]{inputenc} 
\usepackage[T1]{fontenc}    
\usepackage{hyperref}       
\usepackage{url}            
\usepackage{booktabs}       
\usepackage{amsfonts}       
\usepackage{nicefrac}       
\usepackage{microtype}      
\usepackage{lipsum}
\usepackage{fancyhdr}       
\usepackage{graphicx}       
\graphicspath{{media/}}     

\title{SurvCaus : Representation Balancing for Survival Causal Inference}

\author{
  Ayoub Abraich \\
  LaMME \\
  Paris Saclay University / UEVE \\
  Evry, France\\
  \texttt{ayoub.abraich@universite-paris-saclay.fr} \\
   \And
  Agathe Guilloux \\
  LaMME - CNRS\\
  Paris Saclay University / UEVE \\
  Evry, France\\
  \texttt{agathe.guilloux@univ-evry.fr} \\
  \And
  Blaise Hanczar \\
  IBISC \\
  Paris Saclay University / UEVE \\
  Evry, France\\
  \texttt{blaise.hanczar@univ-evry.fr} 
}

\begin{document}
\maketitle

\begin{abstract}
Individual Treatment Effects (ITE) estimation methods have risen in popularity in the last years. Most of the time, individual effects are better presented as Conditional Average Treatment Effects (CATE). Recently, representation balancing techniques have gained considerable momentum in causal inference from observational data, still limited to continuous (and binary) outcomes. However, in numerous pathologies, the outcome of interest is a (possibly censored) survival time.
Our paper proposes theoretical guarantees for a representation balancing framework applied to counterfactual inference in a survival setting using a neural network capable of predicting the factual and counterfactual survival functions (and then the CATE), in the presence of censorship, at the individual level. We also present extensive experiments on synthetic and semisynthetic datasets that show that the proposed extensions outperform baseline methods.
\end{abstract}


\section{Introduction}
\label{intro}
Individual Treatment Effects (ITE) estimation methods have risen in popularity in recent years. These methods often focus on estimating various treatment effects. Most of the time, individual effects are better presented as Conditional Average Treatment Effects (CATE), and the confusion between the two has the potential to hinder progress in personalized research \cite{Brian2021IndividualComparisons}. Conventional methods generally use reweighting or matching approaches to estimate the average treatment effect. Our primary interest is to estimate the CATE at the individual level by estimating each individual's factual and counterfactual survival function. In this paper, we adopt medical terminology, but the methods studied in this work also apply many others domains like economics \cite{Banerjee2009TheEconomics},\cite{SibekoExperimentalAfrica}, politics~\cite{steinberg2004new} or education~\cite{morrison2016searching}.

A randomized clinical trial (RCT) is an ideal way to assess the effect of a treatment on a pathology, according to \cite{Tricoci2009ScientificGuidelines}. In such a trial, the treatment $t=1$ or the placebo $t=0$ is given randomly, i.e., independently of the value of the covariates measured on the individual. This random selection ensures that the covariates in the treated and untreated subpopulations have the same density. In this case, we can use a supervised learning algorithm to measure the effect of the treatment on the outcome of interest, which takes the covariates and treatment as input and our outcome as a label. However, even adequately powered RCTs are not always feasible due to various factors such as cost, time, practical and ethical constraints, and limited generalizability. Most of the time, only data from observational studies are available. 

In an observational study, the choice of treatment is determined by the values of covariates. Consequently, the distributions of the covariates in the treated and untreated subpopulations are different, leading to non-comparability or non-exchangeability, which is a source of confounding bias \cite{Greenland2001ConfoundingIH}. This implies that variations in outcomes between treated and untreated groups could be explained by the treatment, other pre-treatment variables, or both. Therefore, estimating the treatment effect by a supervised algorithm without considering the possible biases will lead to a false estimate.

In numerous pathologies, the outcome of interest is a survival time. So we develop in the present paper a new algorithm for estimating the individual treatment effects with survival outcomes.

\paragraph{Contributions.} Our main contributions are :
\begin{itemize}
    \item A theoretical framework to evaluate and understand representation balancing in causal inference for continuous survival outcomes, in the presence of censoring, with theoretical guarantees. We managed to control the risk of the CATE via a Pinsker-type inequality (see Section \ref{problem}); then, we found a theoretical bound to the counterfactual risk excess by introducing a distance between the factual and counterfactual distributions plunged into a latent space (see Section \ref{derivloss}). 
    \item  A neural network-based method, called SurvCaus, for estimating the factual and counterfactual survival functions at the individual level, and CATE (see Section \ref{survcaus}). 
    \item An empirical study with large-scale experiments that shows SurvCaus outperforms the baseline methods (see Section \ref{experiments}).
\end{itemize}


\section{Related works}
Traditional survival analysis approaches model the treatment effect parametrically by including the treatment as a covariate. The Cox proportional hazards model (CoxPH) \cite{Cox1972RegressionMA} and the accelerated failure time (AFT) model \cite{Wei1992TheAF}, are the most commonly used models, with matching and reweighing techniques. There are causal extensions of the non-parametric Random Survival Forest (RSF) \cite{Ishwaran2008RandomSF} and Bayesian Additive regression trees (Surv-BART) \cite{Sparapani2016NonparametricSA} : RSF applied in a causal survival forest configuration with weighted bootstrap \cite{Cui2020EstimatingHT}; and Surv-BART extended to take into account survival outcomes (Surv-Surv-BART \cite{Sparapani2016NonparametricSA} and AFT-Surv-BART \cite{Henderson2018IndividualizedTE}). For more details, see \cite{Hu2021EstimatingHS}.

It should be noted that these methods do not have a counterfactual prediction mechanism, which is fundamental to the estimation of the Conditional Average Treatment Effects (CATE), defined in literature as the difference between an individual's expected potential outcomes for different treatment conditions. 

Recently, developments in representation learning have made it possible to deal effectively with the problems of high-dimensional data and complex interactions, tho still limited to continuous (and binary) outcomes \cite{Johansson2020GeneralizationEffects}. However, in numerous pathologies, the outcome is measured in terms of survival time in the presence of censure.

Lately, \cite{chapfuwa2020survival} proposed an algorithm to address this issue inspired by \cite{Shalit2017EstimatingIT} by developing a generative model for event times based on planar normalization flows. However, they did not provide theoretical guarantees. 

\section{Problem Statement and Background}
\label{problem}
We begin by introducing the fundamental setup for performing causal survival analysis in observational studies.
\subsection{Notations and context }
We consider $n$ independent individuals. For each individual $i$, $X_i \in \mathcal X \subset \mathbb R^d$ represents its features (context) and  $T_i \in \mathcal T = \{0,1\}$ its binary treatment ($T_i=1$ is usually referred to 'treatment' and $T_i=0$ to 'control'). We also denote by $Y_i$ its survival outcome and $C_i$ its censoring time, such that the observed label is $Y_i^c=Y_i \wedge  C_i$ and $\delta_i =\mathds{1}_{Y_i \leq C_i} $.
For causal reasoning, we need to introduce in addition $Y(t,x)$ and $C(t,x)$ the potential survival and censoring time under treatment $t$ as the feature $x$. The associated potential label is denoted by $Y^c(t,x)=Y(t,x)\wedge C(t,x)$ and $\delta(t,x)=\mathds{1}_{Y(t,x)\leq C(t,x)} $.

Under the STUVA assumption \cite{Rubin2005CausalDecisions} we have that $Y_i^c= Y^c(T_i,X_i)$ and $\delta_i= \delta(T_i,X_i)$. Therefore, our data is noted $\mathcal{D}=\{(X_1,T_1,Y_1^c,\delta_1), \ldots, (X_n,T_n,Y_n^c,\delta_n)\}$ assumed to be i.i.d. from unknown density $p^\star_{(X,T,Y^c,\delta)}$. The marginal density of $X$ is denoted by $p^\star_{X}$, the conditional density of $T | X= x$ by $p^\star_{T | X=x}$, the conditional density of $X | T=t$ by $p^\star_{X | T=t}$. Whenever possible, we will dzop the $x$ dependency, $Y(t,x)=Y(t)$, etc..

Finally, the density of $Y=Y(t)$ conditionally to $T=t,X=x$  (resp. $C(t) \mid T=t, X=x$) is denoted by $f^{\star}_{t}( x,\cdot )$, with c.d.f $F^{\star}_{t}( x,\cdot )$, (resp. $h^{\star}_{t}( x,\cdot )$ with c.d.f $H^{\star}_{t}( x,\cdot )$). 

Throughout this paper, for any cumulative density function (c.d.f.) $G$, $\bar G$ is its associated survival function and $\tau_G = \sup\{t \geq 0, G(y)<1\}.$ The time horizon that we consider is
\begin{equation*}
 \tau_{min}=\min_{(x,t)\in \mathcal{X}\times\{0,1\} }(\tau_{H^\star_t}(x),\tau_{F^\star_t}(x)).   
\end{equation*}

\begin{assumption}\label{ass:ignor_pos}
We assume that $\{Y(0) ,Y(1),C(0),C(1)\} \independent T\ \mid X$ (ignorability) and $\forall (x , t) \in  \mathcal{X} \times \mathcal{T}:$ $0< p^\star_{T | X=x}(t) < 1$ (positivity) \cite{ImbensRecentDI,Pearl2000CausalityMR}. 
\end{assumption}
These assumptions ensure that the CATE is identifiable. However, it is well known that they are not testable in practice. For the ignorability assumption (or equivalently the assumption that they are no unmeasured confounders), we can only hope that the features $X_i$ are sufficiently rich (or in high dimension). The last point makes the positivity assumption less likely to be verifiable (or even verified).

\begin{assumption}\label{ass:indep}
It is further assumed that $Y \independent  C \mid X,T$.
\end{assumption}
 This mechanism is called non-informative censoring \cite{Cole2004AdjustedSC,Daz2019StatisticalIF} and is classical in survival analysis, see e.g. \cite{klein2003survival}. 

\subsection{Problem formulation}\label{sec:problemform}

Our final goal is to estimate the conditional average treatment effect (CATE) that we define, in the context of a survival outcome, as the difference in the respective survival functions at a specific time horizon.   
\begin{definition}\label{def:CATE}
For  $x\in \mathcal{X}$ and hypothesis ($f_{0},f_{1}$), the CATE is defined as follows:
\begin{equation*}
\text{CATE}(f_{0},f_{1},x) = \overline{F}_{1}( x,\tau_{\min} ) -\overline{F}_{0}( x,\tau_{\min} )
\end{equation*}
where $F_{t}( x,\cdot )$ is the c.d.f of $f_t(x,\cdot)$.
\end{definition}
From this definition, one can see that to achieve this goal; a first step is to propose estimates of the unknown densities $f_{0}^{\star},f_{1}^{\star}$ (or their corresponding c.d.f. or survival functions). This CATE has a simple interpretation because, whenever it is positive, the individual will benefit from the treatment in terms of survival probability. It is worth mentioning that different types of CATE  are considered in state-of-the-art, such as differences in expected lifetime or hazard ratio \cite{chapfuwa2020survival}.

The main difficulty in calculating CATE for potential outcome hypotheses is quantifying the counterfactual density (or survival function), which is the focus of this work. Indeed, $Y(t)$ is not observed over the entire population because $Y(1)$ is only observed for treated individuals, and $Y(0)$ is only observed for the control group. Therefore, $f_t^{\star}$ cannot be estimated over the entire population for the same reasons.

The precision of an estimate of the CATE will be measured in terms of the Precision in Estimation of Heterogeneous Effect (PEHE) \cite{Hill2011BayesianInference}, which we now define as the quadratic loss of the CATE.
\begin{definition}
The Precision in Estimation of Heterogeneous Effect denoted by $\text{PEHE}(f_1,f_0) $ of proposals $f_0,f_1$ is defined as follows:
\begin{align*}
\E_{X\sim p^\star_{X}} \Big[\text{CATE}( f_{0} ,f_{1} ,X)  -\text{CATE}( f^{\star}_{0} ,f^{\star}_{1} , X) \Big]^2.
\end{align*}
\end{definition}

The main result of the Section is that the excess risks can bind the PEHE for $f_0$ and $f_1$. To establish these results, we first notice that the definition of our CATE leads to the bound (see Appendix~\ref{app:sec3} for details).
\begin{align*}
\frac18\Big[\text{CATE}( f_{0} ,f_{1} ,x)  -\text{CATE}( f^{\star}_{0} ,f^{\star}_{1} , x) \Big]^2 \leq   \big(d^{x}_{TV}( f_{0} ,f^{\star}_{0})\big)^2  +  \big(d^{x}_{TV}\big( f_{1} ,f^{\star}_{1})\big)^2,
\end{align*} 
where $d_{TV}^{x} $ is the  total variation distance between the densities $f^{\star}_{t}$ and $f_{t}$ at $x$ on $[0,\tau_{min}]$, defined as,
\begin{equation}
\label{dtv}
      d_{TV}^{x} \big( f^{\star}_{t} ,f_{t}\big)=\frac{1}{2}\int _{0}^{\tau_{min}}| f_{t}( x,y) \ -f_{t}^{\star}( x,y)| dy.
\end{equation}

Define, for $x\in\mathcal{X}$,  the expected point-wise loss $\ell _{f_t}( x)$ is for a hypothesis $f_t \in \mathcal{H}$ as
\begin{equation*}
  \ell _{f_t}( x) : =\mathbb{E}_{(Y^{c}(t) ,\delta(t) ) \mid X}\Big[ L\Big( X,Y^{c}(t) \ ,\delta(t),f_t\Big)\mid X=x\Big]
\end{equation*} 
where $L$ is the negative log-likelihood for survival data (see Section \ref{app:sec3} of Appendix). Associated to this loss, we define the Kullback-Leibler divergence as
\begin{equation}\label{eqn:kullback}
   \KL_x \Big( f^{\star}_{t} ||f_{t}\Big)  = \ell _{f_{t}}( x) -\ell _{f^{\star}_{t}}( x).
\end{equation}

Now, with the use of a particular Pinsker type inequality~\cite{tsybakov2003introduction} (see Appendix~\ref{app:sec3} for a proof) bounding the total-variation by the Kullback-Leibler divergence, we obtain the bound
\begin{align*}
&\frac14\Big[\text{CATE}( f_{0} ,f_{1} ,x)  -\text{CATE}( f^{\star}_{0} ,f^{\star}_{1} , x) \Big]^2\leq
\frac{1}{ \eta^2 } \Big(  \KL_x \big( f^{\star}_{0} ||f_{0}\big)+  \KL_x \big( f^{\star}_{1} ||f_{1}\big)\Big).
\end{align*}
where
\begin{equation*}
\label{eta}
   0<\eta \leq \min_{(x,t)\in \mathcal{X}\times\{0,1\} } \overline{H}^{\star}_{t}( x,\tau _{}),
\end{equation*}
for any $\tau<\tau_{min}$. 

Now, we define the marginal risk as  of a hypothesis $f_t$ as
\begin{equation*}
    R(f_t) =\E_{X \sim p^\star_X} \ [ \ell _{f_t}( X)].
\end{equation*} and the excess risk as
\begin{equation}
\label{erdef}
\ER( f _{t})=R( f _{t}) -R\big( f^{\star}_{t}\big) =\E_{X\sim p^\star_X}\big[ \KL_{X}\big( f ^{\star}_{t} ||f _{t}\big)\big].
\end{equation}
We can now state the main result of this section.
\begin{theorem} [Bound risk for the PEHE]
\label{theo:ctrlpehe}
For any hypothesis $(f_{0},f_{1})$, the PEHE verifies
\begin{equation} \label{eqn:ctrlpehe}
         \text{PEHE}(f_1,f_0) \leq  \frac{4}{\eta^2 } \{\ER( f _{0}) + \ER( f _{1})\}.
\end{equation}
\end{theorem}
This shows that small values of the excess risks for the $(f_0,f_1)$ hypothesis guarantee a small PEHE. In other words, if we estimate well $f^\star_0,f^\star_1$, we guarantee a good estimate of the CATE.
Details for this Section can be found in Section~\ref{app:sec3} of Appendix.

\section{Bounding the Excess Risks}
\label{sec:bounds}

As the excess risks  $\ER(f_0)$ and $\ER(f_1)$ are not directly estimable because they involve the distributions of  counterfactual quantities, we propose in this Section to bound them by quantities that can be easily estimated from the factual data.
\subsection{Importance-reweighing}
Towards that end, we will now consider weights and introduce the factual (resp. counterfactual) weighted excess risk. 
\begin{definition}
For weighting function $w : \mathcal{X}\times \mathcal{T} \to \mathbb{R}^{+}$, satisfies for all $t\in\mathcal{T}$
\begin{equation*}
    \E_{X\mid T \sim p^\star_{X | T=t}}[ w(X,T) \mid T=t] =1.
\end{equation*}
We define as $\ER_t^w(f_t)$ (resp. $ \ER_{1-t}^w(f_t)$) the factual (resp. counterfactual) weighted excess risk \cite{Shimodaira2000ImprovingPI}, defined as 
\begin{align*}
     &\ER_b^w(f_t) =  \E_{X\mid T \sim p^{\star,w}_{X | T=b}}\Big[  \ell _{f_{t}}(X) -\ell _{f^{\star}_{t}}( X)\mid T=b\Big] \\
     &=\E_{X\mid T \sim p^\star_{X | T=b}}\Big[ w(X,T)\KL_{X}\Big( f ^{\star}_{t} ||f _{t}\Big)\mid T=b\Big]
\end{align*}
for $ b\in \{ t,1-t\}$, where the factual weighted conditional density of $X \mid T=t$ (resp. counterfactual weighted conditional density $X \mid T=1-t$)  are defined as $ p^{\star,w}_{X | T=t}( x) =w(x,t) p^\star_{X | T=t}(x)$ (resp. $ p^{\star,w}_{X | T=1-t}( x) =w(x,t) p^\star_{X | T=1-t}(x)$ ).
\end{definition}
We denote $ \ER_t(f_t)=\ER_t^{w=1}(f_t)$ (resp. $  \ER_{1-t}(f_t)=\ER_{1-t}^{w=1}(f_t)$) the factual (resp. counterfactual) excess risk. The treatment group is indicated by the index $t$ on excess risk $\ER_t$. It is important to note that the potential outcome against which the excess risk is evaluated is implied in this notation.
The factual excess risk $\ER_t(f_t) $ is estimable under ignorability, it's also in general a biased estimator of $\ER(f_t)$ in general, which is not directly estimable because
\begin{equation}\label{eqn:decomp_ER}
     \ER(f_t) =  \alpha_t\underbrace{  \ER_t(f_t) }_{\text{estimable}}+ (1-\alpha_t)  \underbrace{\ER_{1-t}(f_t)}_{\text{non-estimable}}
\end{equation}
where $\alpha _{t} =\mathbb P ( T=t)$, which will have a strong impact on the estimation of $f_t^\star$ and the $\CATE$. See Appendix \ref{app:section4} for a proof.
In what follows, we somehow follow the main steps as in \cite{johansson2020generalization}, but it, however, is worth mentioning that they are significant differences: 
i) we focus on excess risk instead of marginal risk; ii) we do not consider the square loss. 

Going back to Equation~\eqref{eqn:decomp_ER}, to bound the risk $\ER(f_t)$ of $f_{t}$ on the whole population,  we first rewrite it, see Appendix~\ref{app:section4} for details.
\begin{lemma} \label{lemma:decom}Defining $\tilde{w}( x,t) =\alpha _{t} +( 1-\alpha _{t})w(x,t)$, we have
\begin{equation*}
    \ER( f _{t}) =\underbrace{ \ER^{\tilde{w}}_{t}( f_{t}) }_{\text{estimable}}+ \alpha _{1-t}\underbrace{\Big[ \ER_{1-t}( f_{t}) -\ER^{w}_{t}( f _{t})\Big]}_{\Delta_{t}^w( f_{t})}.
\end{equation*}
\end{lemma}
This brings us closer to a bound for the PEHE. We indeed exhibit, in the next section, a bound for $\Delta_{t}^w( f_{t})$. We first introduce some notations related to balanced representation learning and assumptions that will serve us in the following.
\subsection{Balanced representation learning}

Let $\mathcal{E} \subset\{\mathcal{X} \to \mathcal{Z}\}$ denote a family of representation functions of the contexts space into a latent space $\mathcal{Z}$. A $\phi \in \mathcal{E}$ is called an embedding function. Further, let $\mathcal{G} \subseteq\{h: \mathcal{Z} \times \mathcal{Y} \times \mathcal{T} \to \mathbb{R}^+ \}$ denote a set of hypotheses and let $\mathcal{H}$ be the space of all such compositions
\begin{equation*}
    \mathcal{H}=\{f^\phi(\cdot,\cdot,\cdot)=h(\phi(\cdot),\cdot,\cdot): h \in \mathcal{G}, \phi \in \mathcal{E}\}.
\end{equation*}
We consider learning $\phi$ while minimizing the excess risk of hypotheses $f_t^\phi(\cdot,\cdot)=f^\phi(\cdot,\cdot,t)=h(\phi(\cdot),\cdot,t)\in \mathcal{H}$ for $t=0,1$ (see the objective loss defined in Section~\ref{eqn:obj_loss}).

For the CATE to be estimable from the factual data, we precisely need the same assumptions (see \ref{ass:ignor_pos}) on $\phi(X)$ as previously on $X$  (see \cite{johansson2020generalization}).
\begin{assumption}\label{ass:ignor_pos_phi}
We assume that $\{Y(0) ,Y(1),C(0),C(1)\} \independent T\ \mid \phi(X)$ (ignorability) and $\forall (z, t) \in  \mathcal{Z} \times \mathcal{T}:$ $0< p^\star_{T | \phi(X)=z}(t) < 1$ (positivity).
\end{assumption}

It is impossible to verify the assumptions \ref{ass:ignor_pos_phi} for a given $\phi$ based uniquely on factual data. To solve this, we consider learning twice-differentiable, invertible representations $\phi: \mathcal{X} \to \mathcal{Z}$ where $\Psi: \mathcal{Z} \to \mathcal{X}$ is the inverse representation, such $\psi=\phi^{-1}$. The invertibility of $\phi$ with assumptions \ref{ass:ignor_pos}  on $X$  implies the assumptions \ref{ass:ignor_pos_phi} on $\phi(X)$. So we drop this hypothesis, keeping only the hypotheses \ref{ass:ignor_pos},  and we obtain the following result.

\begin{theorem}
\label{thm:bound}
Keeping the previous notation with $p^{\star,\phi,w}_{X | T=t}(x)=p^{\star,w}_{X | T=t}(\psi(x))$ and under certain conditions (see Appendix~\ref{app:balancedproof}), there is a constant $C_{\phi }>0$ such that,
\begin{equation*}
    \Delta _{t}^w( f_{t}^\phi) \leq C_{\phi }  \text{IPM}_{\mathcal{L}}\Big(  p^{\star,\phi}_{X | T=1-t} ,p^{\star,\phi,w}_{X | T=t}\Big)
\end{equation*}
 Therefore,
\begin{equation*}
\ER( f _{t}^\phi) \leq \ER^{\tilde{w}}_{t}(f _{t}^\phi) +\alpha _{1-t}  C_{\phi }  \text{IPM}_{\mathcal{L}}\Big(  p^{\star,\phi}_{X | T=1-t} ,p^{\star,\phi,w}_{X | T=t}\Big)
\end{equation*}
where the Integral Probability Metrics (IPM) \cite{Mller1997IntegralPM} is defined as 
\begin{equation*}
    \text{IPM}_{\mathcal{L}}(p,q)=\sup _{g\in \mathcal{L}}\Big| \mathbb{E}_{X\sim p}[ g( X)] -\mathbb{E}_{X \sim q}[ g( X)]  \Big|
\end{equation*} 
and  $\mathcal{L}$ is a reproducing kernel Hilbert space (RKHS) induced by a universal kernel \cite{Gretton2012ATest}.
\end{theorem}
\begin{remark}
If $\mathcal{L}$ is the set of functions of norm 1 in an RKHS, the IPM is Maximum Mean Discrepancy (MMD) \cite{Gretton2012ATest}. If $\mathcal{L}$ is the set of Lipschitz functions of the norm at most 1, the IPM becomes the Wasserstein distance \cite{Villani2007OptimalNew}, which we will adopt in our algorithm for various reasons such as improving learning stability, getting rid of problems like mode collapse, see \cite{Arjovsky2017WassersteinG, Pinetz2019OnTE}.
\end{remark}
Combining the previous elements and denoting $p^{\star,\phi,w_t}_t=p^{\star,\phi,w_t}_{X | T=t}$, we just established that the PEHE (times $\eta^2/(4\beta)$) is bounded by
\begin{equation}\label{eq:boundExpectPEHE}
    R^{\tilde{w}}(f^\phi)   +  \frac{C_{\phi}}{\beta}   \text{IPM}_{\mathcal{L}}\big(  p^{\star,\phi,w_0}_{0}, p^{\star,\phi,w_1}_{1}\big)
\end{equation}
plus a term that does not depend on $f^\phi$ and
 where $R^{\tilde{w}}(f^\phi)$ is the weighted factual risk integrated over the distribution $p^\star_{(X,T,Y^c,\delta)}$, see a detailed definition and proof in Appendix~\ref{app:balancedproof}.

\subsection{Derivation of our loss}\label{derivloss}
The derivation of our loss comes from the bounding of the two terms of Equation~\eqref{eq:boundExpectPEHE} by their empirical counterparts. We give in this paragraph the main arguments to derive such a bound to explain the rationale behind our loss.

Let define the empirical weighted risk as
 \begin{equation*}
   \widehat{R}^{\tilde{w}}(f^\phi)=  \frac{1}{n}\sum ^{n}_{i=1} \tilde{w}( \phi(x_{i}) ,t_{i})L\Big(x_i, y_i, \delta_i, f^\phi\Big).
 \end{equation*}
According to classical results of statistical theory theory, see \cite{vapnik1999nature,Cortes2010LearningWeighting}, under certain moment conditions, we have with a high probability
\begin{equation*}
    R^{\tilde{w}}(f^\phi) \leq \widehat{R}^{\tilde{w}}(f^\phi) + \mathcal{O}(\frac{1}{n^{3/8}}).
\end{equation*}
From \cite{Sriperumbudur2009OnIP}, we know that, with high probability
\begin{equation*}
\text{IPM}_{\mathcal{L}}\Big(  p^{\star,\phi,w_0}_{0}, p^{\star,\phi,w_1}_{1}\Big) \leq \text{IPM}_{\mathcal{L}}\Big(\hat{p}^{\phi,w_{1}}_{1} ,\hat{p}^{\phi,w_{0}}_{0}\Big) + \mathcal{O}(\frac{1}{n^{1/2}})
\end{equation*}
where $\hat{p}^{\phi,w_{t}}_{t}$ is the empirical distribution associated to $p^{\star\phi,w_{t}}_{t}$, we refer the readers to Appendix~\ref{app:section4} for proper definitions.

Following the two last results, we know that, with high probability, the PEHE (times $\eta^2/(4\beta)$) is bounded by
\begin{align*}
 \widehat{R}^{\tilde{w}}(f^\phi) + \frac{C_{\phi}}{\beta} \text{IPM}_{\mathcal{L}}\Big(\hat{p}^{\phi,w_{1}}_{1} ,\hat{p}^{\phi,w_{0}}_{0}\Big) +\mathcal{O}(\frac{1}{n^{3/8}})   
\end{align*}plus a term that does not depend on $f^\phi$. This justifies our choice for the loss, in which we finally add two regularization terms
\begin{align*} \label{eqn:obj_loss}
    \mathcal{\tilde{O}}(f^\phi,\phi,w,\lambda_r,\lambda_w,\gamma_{\text{wd}})=\underbrace{\sum_{i=1}^n \frac{\tilde{w_{i}}}{n} L(y_i,x_i,t_i,\delta_i,f^\phi) }_{\hat{R}^{\tilde{w}}(f^\phi)} \frac{\gamma_{\text{wd}}}{n}  \underbrace{\text{IPM}_{\mathcal{L}}\Big(\hat{p}^{\phi,w_{1}}_{1} ,\hat{p}^{\phi,w_{0}}_{0}\Big)}_{\text{Distributional distance}}  + \underbrace{\frac{\lambda_r}{\sqrt{n}}\Omega(f^\phi)+ \frac{\lambda_w}{n} \Theta(w) }_{\text{Regularization}} 
\end{align*}
where $\tilde{w_{i}}=\tilde{w}(\phi(x_i),t_i)$. 

\section{SurvCaus Netwrok}
\label{survcaus}
SurvCaus is a deep learning architecture that has been tuned to estimate survival functions for a continuous time of relapse in the presence of censoring, over the interval $[0, \tau_{min}]$ and CATE at the individual level by aligning factual and counterfactual distributions over a representation space.

\paragraph{Discretization of Durations} 
For our method to work on a continuous time data, a discretization of time is required in the form $0=\tau _{0} < \tau _{1} < \dotsc < \tau _{m} =\tau_{min}$. In addition, for intrinsically discrete event times, we may want to minimize $m$ discrete timescale, as this reduces the number of parameters in the neural networks. The most obvious method for discretizing time is to create an equidistant grid of m grid points. Another approach, explored in \cite{kvamme2019continuous}, is to create a grid based on the density of event times by estimating the survival function $ =\hat{S}_{KM}(t)$ with the Kaplan-Meier estimator. Let $ 0< \eta _{i} -\eta _{i+1} = (1-\eta _{m} )(m)$ such as $ \tau _{i} \ =\hat{S}^{-1}_{KM} \ ( \eta _{i}) \ $  for $ i=1,\dotsc ,m$.

We denote $\text{Sub}( \tau_{min} ,m) =\{\tau _{1} ,\cdots ,\tau _{m}\}$ and $ k(y) \in \{1, ,m\}$ the index, such as $ y\in I_{k(y)}$. It is assumed that the density $f_t(x,\cdot)$ is piecewise constant over each $ I_{i}$, with $f_t(x,y)=f_t(x,\tau_{k(y)}) $.

\paragraph{Model output}
Let $\phi$ and $\Psi$ two multilayer neural networks such as $\Psi ( x,t) =[\Psi_{1}( x,t) ,\dotsc ,\Psi_{m}( x,t)] \in \mathbb{R}^{m}$, with $m$ the subdivision slope $\text{Sub}( \tau_{min} ,m)$,  such as, the output of our network is,
\begin{equation*}
    f^{\phi }_{t}( x ,\tau _{k}) =\frac{\exp[ \Psi_{k}( \phi ( x) ,t)]}{1+\sum ^{m}_{j=1}\exp[ \Psi_{j}( \phi ( x) ,t)]} =\ \sigma ^{t}_{k}( \Psi ,\phi ,x),
\end{equation*}
see \cite{lee2018deephit} or \cite{kvamme2019continuous} for similar architectures.

\paragraph{Survival functions}
Under the assumption that the output of our network is a density (i.e. with sum equal to 1), we require the condition $ \sigma ^{t}_{m+1}( \Psi ,\phi ,x) = (1+\sum ^{m}_{j=1}\exp[ \Psi_{j}( \phi ( x) ,t)])^{-1}$, that corresponds to $\Psi_{m+1}=0$.
So we get the survival functions prediction as 
\begin{equation*}
   \widehat{\overline{F_{t}}}( x,y) =\sum ^{m+1}_{j=k(y)+1} \sigma ^{t}_{j}( \Psi ,\phi ,x)  
\end{equation*}

\paragraph{Loss function parameterization}
We now specify the terms that appear in our objective loss $ \mathcal{\tilde{O}}$. The survival loss (see Equation~\eqref{eq:survloss}) after discretization and soft-max parametrization writes
\begin{align*}
    &L(y_i,x_i,t_i,\delta_i,f^\phi)=L(y_i,x_i,t_i,\delta_i,\Psi ,\phi)=- \delta _{i}\log \sigma ^{t_i}_{k(y_i)}( \Psi ,\phi ,x_i) - ( 1-\delta _{i})  \log \sum ^{m+1}_{j={k(y_i)}+1} \sigma ^{t}_{j}( \Psi ,\phi ,x_i).
\end{align*}

We choose to regularize our loss by ridge penalties, so we set
\begin{align*}
    \Omega(f^\phi)=  \| \Psi \|_2; \;\Theta(w)=\| w \|_2.
\end{align*}
Finally the distributional distance $\text{IPM}_{\mathcal{L}}$ is taken as the Wasserstein distance $ d_{\text{WD}}$ and is computed using Sinkhorn's algorithm, see \cite{Cuturi2013SinkhornDL}.

\section{ Experiments}
\label{experiments}
\subsection{Prediction task and benchmark}

\paragraph{Interpolation for Continuous-Time Predictions}

As a result of our discretization, the survival estimates become a step function with steps at grid points. Therefore, it may be advantageous for coarser grids to interpolate the discrete survival estimates. Inspired by \cite{kvamme2019continuous}, we interpolate with a simple linear scheme that meets the monotonicity requirement of the survival function. Our model performs better with this interpolation than interpolating the survival function as a piecewise constant (see section \ref{app:experiments} in Appendix).

\paragraph{Evaluation scores}\label{evalscore}
To evaluate the performances of our algorithm and its competitors, we define the following metrics:
\begin{align*}
  \text{MiseSurv}^2(x,t)&= \big\|\bar{F}_t^{\star}(x,\cdot) -  \widehat{\overline{F_{t}}}(x,\cdot) \big\|_{[0,\tau_{\min}]}^2
  \\ \text{MiseCate}^2(x)&=\big\|\CATE^{\star}(x,\cdot) - \widehat{\CATE}(x,\cdot) \big\|_{[0,\tau_{\min}]}^2 \\ &\leq 2\big ( \text{MiseSurv}(x,0 )^2 + \text{MiseSurv}(x,1 ) ^2 \big ) =  2\text{FSMise}^2(x),
\end{align*}
their means $\text{MCATE}, \text{FSM}$ over the test dataset and 
\begin{equation*}
    \text{MPEHE}= \frac{1}{nm}\sum_{i,j}^{}\big|\CATE^{\star}(x_i,\tau_j) - \widehat{\CATE}(x_i,\tau_j) \big|^2. 
\end{equation*}

\paragraph{Benchmark and validation}

Predictive performances of SurvCaus Network in predicting the CATE are compared in terms of PEHE, MCATE, and FSM, with five baseline methods:  Surv-BART \cite{Sparapani2016NonparametricSA} form R library \href{https://rdrr.io/cran/BART/man/surv.bart.html}{\texttt{surv.Surv-BART}} and  CoxPH \cite{Buchanan2014WorthTW}, DeepSurv \cite{Katzman2018DeepSurvPT}, EST \cite{Geurts2006ExtremelyRT} and RSF \cite{Ishwaran2008RandomSF} from \href{https://square.github.io/pysurvival/}{\texttt{PySurvival}} library. 

SurvCaus is trained on the entire training data set, whereas state-of-the-art models are trained on the subsets of treated and untreated patients in the training dataset separately, as training them on the entire data set produces erroneous estimates. SurvCaus is implemented in Python in a Pytorch environment. $\phi$ and $\Psi$ implemented in 4 layers with 221 ReLU neurons, Xavier Gaussian initialization
  schemes, Adam optimizer, 256 examples per mini-batch, and early stopping. The hyperparameters include the number of subdivisions $N_{\text{durations}}$, the learning rate, the regularization penalty parameters $\lambda_r,\lambda_w,\gamma_{\text{wd}}$. The SurvCaus hyperparameters and those of the competing benchmark models are optimized using random search \cite{Bergstra2012RandomSF}. For each hyperparameter, we set a discrete search space using manual search. The performance of the models is then calculated on a bootstrap of 50 experiments. 

\subsection{Data set}

Our experiments are performed on both synthetic and real datasets that we describe in the following. Table \ref{data}) shows the main characteristics of these datasets.

\paragraph{Synthetic data}
The generation of our <synthetic datasets follows the algorithm below. For a sample size $n$ and $p$ features, and for each individual $i=1, \ldots,n$, we first simulate its features $x_i$ according to the multivariate Gaussian $\mathcal{N}_{n,p}(\vec{0}_p,\Sigma)$ where $\Sigma$ is a Toeplitz matrix of size $n \times p$ and $\rho=0.1$. The treatment $t_i$ of individual $i$ is then chosen according to a binomial distribution of parameter $p_{i}$ where 
 \begin{equation*}
     p_{i} =\text{sigmoid}((-1)^i \exp(i/10)).
 \end{equation*}
Then, to control the distance between the distribution of the features among treated individuals and untreated ones, we transform the features via the translation $x_i \gets x_i+ p_{wd} \times (2t_i-1)$
where $p_{wd}$ is a parameter that controls the Wasserstein distance. We then simulate the factual and counterfactual survival times $ Y(t)$  according to the survival functions $ \overline{F}_{t}^{\star}$ ($t=0,1$) defined as
\begin{equation*}
 \overline{F}_{t}^{\star}(x,y) =\exp\Big[ -( \lambda y)^{\alpha}\exp( s(x) \ +\ \epsilon t)\Big]   
\end{equation*}
where $\alpha=2$ and $\epsilon=1.8$ are fixed.
We consider two different simulation scenarios: a linear scheme  (LS) and a nonlinear scheme (NLS), see Appendix~\ref{app:experiments} for more details. The censoring times are simulated from an exponential distribution $\mathcal E(\lambda_c)$ where $\lambda_c$ is chosen to achieve a censoring of about 30\%. 

It should be noted that the choice of simulation parameters is made in order to have a regularity on the survival time for both treated and untreated groups, i.e. to have a time range that covers the factual and counterfactual time $\tau_{min}=\min(\tau_{H_0}^\star,\tau_{H_1}^\star,\tau_{F_0}^\star,\tau_{F_1}^\star)$, which is not always true, but is necessary for our theoretical framework. With this simulation scheme, we create  train, test and validation datasets of (60\%,20\%,20\%) proportions respectively. We denote $d_{\text{WD}}^{\text{init}}=d_{\text{WD}}(\hat{p}_{1} ,\hat{p}_{0})$ the Wasserstein distance on initial space $\mathcal{X}$.

\begin{figure}[ht]
 \caption{ \% FSM in function of $\gamma_{\text{wd}}$ on synthetic dataset }
  \label{alphawd FSM}
  
 \begin{center}
\centerline{\includegraphics[width=\columnwidth]{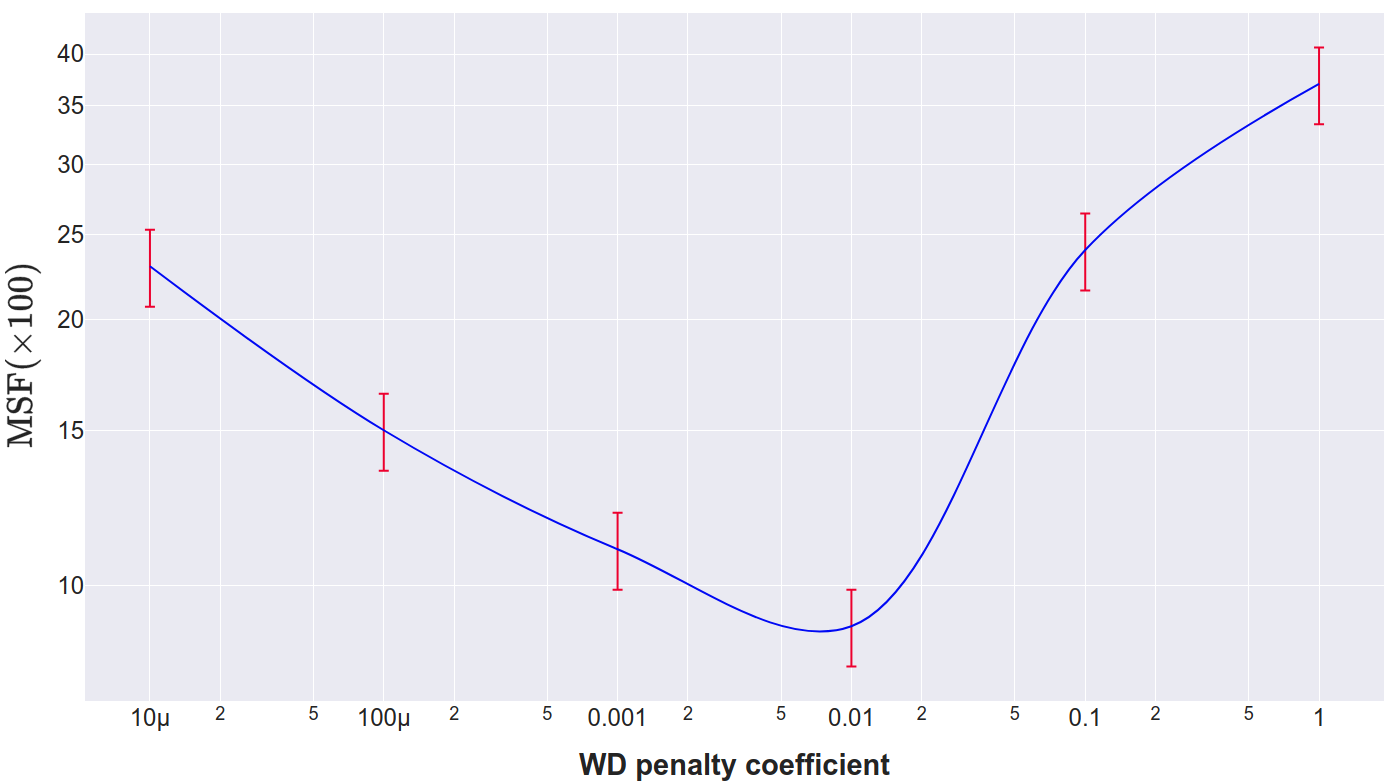}}
\label{fig:alpha}
\end{center}
\vskip -0.3in
\end{figure}

\begin{figure}[ht]
 \caption{ LS : FSM in function of $d_{\text{WD}}^{\text{init}}$ on synthetic dataset}
 \label{LS FSM}
 \begin{center}
\centerline{\includegraphics[width=\columnwidth]{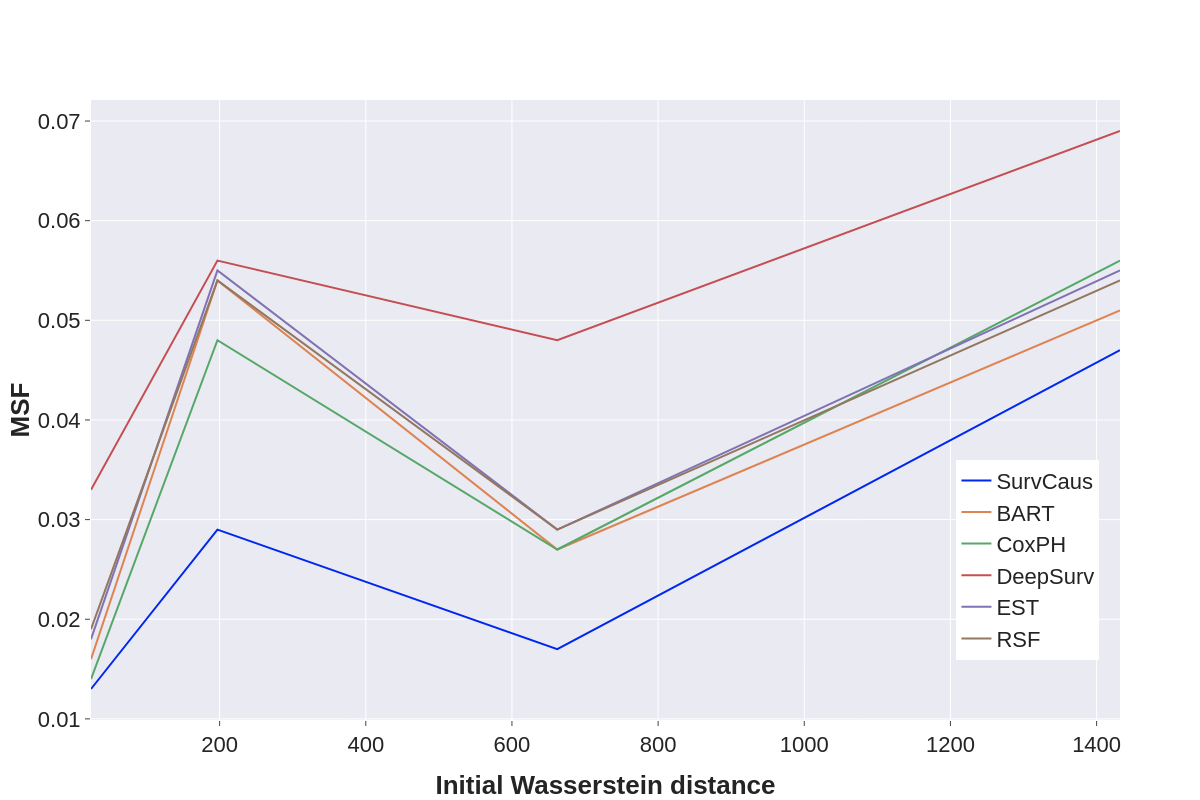}}
\label{fig:LS}
\end{center}
\vskip -0.3in
\end{figure}

\begin{figure}[ht]
 \caption{ NLS : FSM in function of $d_{\text{WD}}^{\text{init}}$ on synthetic dataset }
  \label{NLS FSM}
 \begin{center}
\centerline{\includegraphics[width=\columnwidth]{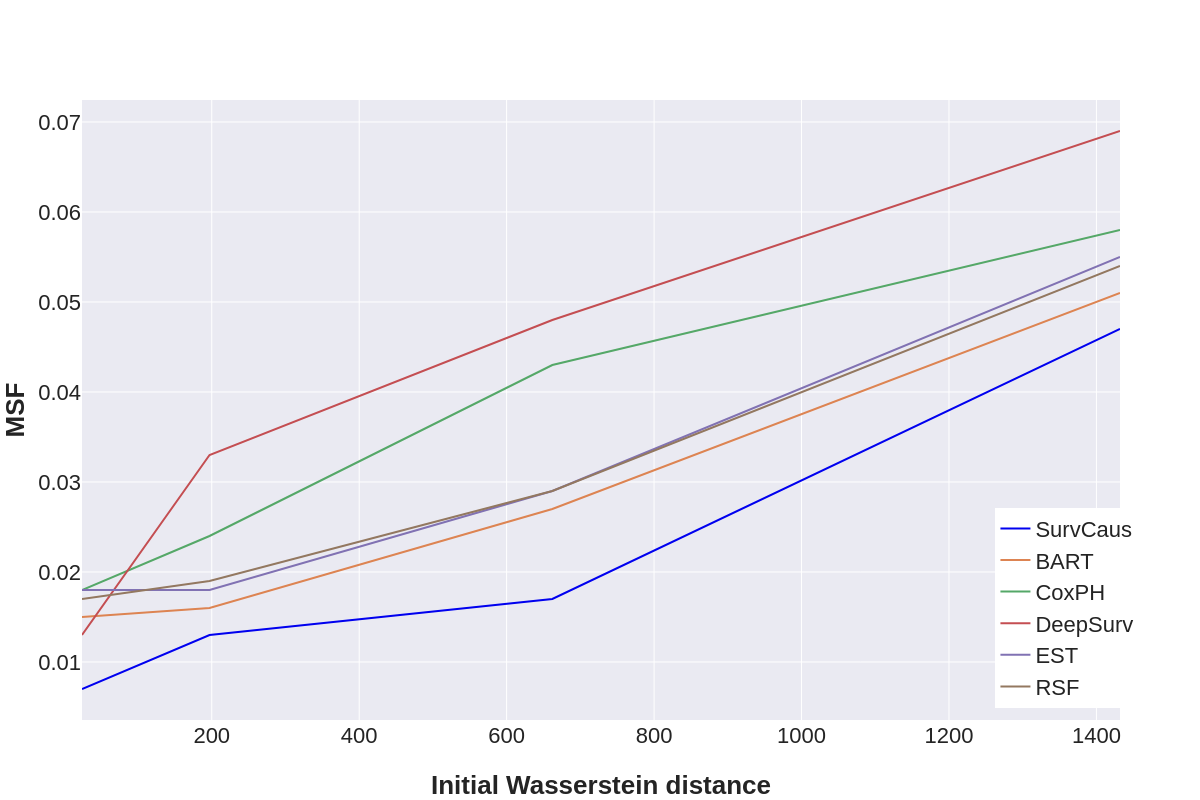}}
\label{fig:NLS}
\end{center}
\vskip -0.3in
\end{figure}

\paragraph{Real data}

We run experiments on real data sets : i) RNA-Seq from The Cancer Genome Atlas Program (TCGA) \cite{Weinstein2013TheCG}; ii) Study to Understand Prognoses Preferences Outcomes and Risks of Treatment (SUPPORT) \cite{1990SUPPORTST}) ; iii)  Molecular Taxonomy of Breast Cancer International Consortium (METABRIC) \cite{Curtis2012TheGA}.

The datasets are available in the the \href{https://github.com/havakv/pycox}{\texttt{Pycox} python package}, see \cite{kvamme2019continuous}, and require no additional preprocessing. Since couterfactual outcomes are not available for real data, we simulated outcomes with the same schemes as described above. We created train, test and validation sets of (60\%,20\%,20\%) proportions respectively.

\begin{table}[ht]\centering
\caption{Summary of the three data sets used in the experiments.}\label{data}
\begin{tabular}{lrrrr}\toprule \textbf{Dataset}
&\textbf{Size} &\textbf{$N_{\text{features}}$} &\textbf{Prop. Censored} \\\midrule
\textbf{SUPPORT} &8 873 &14 &0.32 \\
\textbf{METABRIC} &1904 &9 &0.42 \\
\textbf{TCGA} &953 &221 &0.31 \\
\textbf{Synthetic} &1000 & 35 & 0.30 \\
\bottomrule
\end{tabular}
\end{table}

\subsection{Results}

\begin{table*}[ht]
\caption{Predictive performance (mean $\pm$ sd) on the validation sets (internal validation) for the linear simulated data set. In the results, $\dagger$ italicized indicates the performance difference between SurvCaus and the method from the state-of-the-art is insignificant (i.e. p-value greater than 0.05). For simplicity of notation, significant results of p-value $< 0.05$ are not marked. }
\label{table:linear}
\vskip 0.15in
\begin{center}
\begin{small}
\begin{sc}
\resizebox{1\columnwidth}{!}{
\begin{tabular}{lrrr|rrr|rrr|rrrr}\toprule
&\multicolumn{3}{c}{Synthetic data } &\multicolumn{3}{c}{TCGA } &\multicolumn{3}{c}{SUPPORT} &\multicolumn{3}{c}{METABRIC} \\\cmidrule{2-13}
\textbf{} &\textbf{MCATE} &\textbf{MPEHE} &\textbf{FSM} &\textbf{MCATE} &\textbf{MPEHE} &\textbf{FSM} &\textbf{MCATE} &\textbf{MPEHE} &\textbf{FSM} &\textbf{MCATE} &\textbf{MPEHE} &\textbf{FSM} \\\midrule
\textbf{SurvCaus (ours)} &\textbf{0.09$\pm$0.04} &\textbf{0.16$\pm$ 0.05} &\textbf{0.05$\pm$0.05 } &\textbf{0.04$\pm$0.02} &\textbf{0.29$\pm$0.05} &\textbf{0.02$\pm$0.01} &\textbf{0.03$\pm$0.01} &\textbf{0.06$\pm$0.03} &\textbf{0.01$\pm$0.01} &\textbf{0.01$\pm$0.01} &\textbf{0.03$\pm$0.01} &\textbf{0.01$\pm$0.01} \\
Surv-BART &0.16$\pm$ 0.05 &0.26$\pm$0.05 &0.07$\pm$ 0.03 &0.08$\pm$0.01 &0.43$\pm$0.16 &0.05$\pm$0.01 &0.05$\pm$0.01 &0.08$\pm$0.06 &\emph{0.02$\pm$0.01$^\dagger$} &0.02$\pm$0.01 &0.04$\pm$0.02 &\emph{0.02$\pm$0.01$^\dagger$}\\
CoxPH &0.32$\pm$ 0.11 &0.54$\pm$0.08 &0.18$\pm$ 0.1 &0.08$\pm$0.04 &0.47$\pm$0.13 &0.04$\pm$0.02&0.09$\pm$0.03 &0.15$\pm$0.10 &0.04$\pm$0.01 &0.03$\pm$0.01 &0.06$\pm$0.01 &0.03$\pm$0.02 \\
DeepSurv &0.29$\pm$ 0.11 &0.51$\pm$0.03 &0.29$\pm$0.19 &0.08$\pm$0.05 &0.5$\pm$0.17 &0.05$\pm$0.03 &0.14$\pm$0.06 &0.20$\pm$0.20 &0.05$\pm$0.03 &0.03$\pm$0.03 &0.06$\pm$0.03 &0.03$\pm$0.01 \\
EST &0.17$\pm$0.03 &0.27$\pm$ 0.06 &0.09$\pm$0.03 &0.09$\pm$0.02 &0.46$\pm$0.14 &0.05$\pm$0.01 &0.04$\pm$0.01 &0.07$\pm$0.04 &\emph{0.02$\pm$0.01$^\dagger$} &0.02$\pm$0.01 &0.04$\pm$0.01 &0.03$\pm$0.02 \\
RSF &0.15$\pm$ 0.04 &0.25$\pm$0.05 &\emph{0.08$\pm$0.03$^\dagger$} &0.09$\pm$0.02 &0.45$\pm$0.14 &0.05$\pm$0.01 &0.05$\pm$0.02 &0.09$\pm$0.05 &\emph{0.02$\pm$0.01$^\dagger$} &0.03$\pm$0.01 &0.05$\pm$0.01 &0.03$\pm$0.01 \\
\bottomrule
\end{tabular}
}
\end{sc}
\end{small}
\end{center}
\vskip -0.1in
\end{table*}

\begin{table*}[ht]
\caption{Predictive performance (mean $\pm$ sd) on the validation sets (internal validation) for the non-linear simulated data set}
\label{table:nonlinear}

\vskip 0.15in
\begin{center}
\begin{small}
\begin{sc}
\resizebox{1\columnwidth}{!}{%
\begin{tabular}{lrrr|rrr|rrr|rrrr}\toprule
&\multicolumn{3}{c}{Synthetic data } &\multicolumn{3}{c}{TCGA } &\multicolumn{3}{c}{SUPPORT} &\multicolumn{3}{c}{METABRIC} \\\cmidrule{2-13}
&\textbf{MCATE} &\textbf{MPEHE} &\textbf{FSM} &\textbf{MCATE} &\textbf{MPEHE} &\textbf{FSM} &\textbf{MCATE} &\textbf{MPEHE} &\textbf{FSM} &\textbf{MCATE} &\textbf{MPEHE} &\textbf{FSM} \\\midrule
\textbf{SurvCaus (ours)} &\textbf{0.007$\pm$0.004} &\textbf{0.034$\pm$ 0.014} &\textbf{0.007$\pm$0.003} &\textbf{0.03$\pm$0.02} &\textbf{0.31$\pm$0.05} &\textbf{0.03$\pm$0.01} &\textbf{0.06$\pm$0.01} &\textbf{0.06$\pm$0.03} &\textbf{0.05$\pm$0.01} &\textbf{0.02$\pm$0.01} &\textbf{0.03$\pm$0.01} &\textbf{0.02$\pm$0.01} \\
Surv-BART &0.011$\pm$ 0.004 &0.051$\pm$ 0.024 &0.014$\pm$0.003 &0.09$\pm$0.01 &0.43$\pm$0.16 &0.08$\pm$0.01 &0.09$\pm$0.01 &0.11$\pm$0.06 &0.08$\pm$0.02 &0.03$\pm$0.01 &0.04$\pm$0.02 &0.03$\pm$0.01 \\
CoxPH &0.089$\pm$ 0.016 &0.323$\pm$0.259 &0.052$\pm$ 0.009 &0.18$\pm$0.09 &0.47$\pm$0.13 &0.17$\pm$0.02 &0.21$\pm$0.03 &0.25$\pm$0.10 &0.23$\pm$0.03 &0.03$\pm$0.01 &0.04$\pm$0.01 &0.04$\pm$0.01 \\
DeepSurv &0.099$\pm$0.019 &0.364$\pm$0.288 &0.064$\pm$ 0.011 &0.17$\pm$0.05 &0.51$\pm$0.17 &0.18$\pm$0.03 &0.14$\pm$0.06 &0.20$\pm$0.20 &0.15$\pm$0.03 &0.05$\pm$0.03 &0.08$\pm$0.03 &0.06$\pm$0.01 \\
EST &0.011$\pm$0.004 &0.053$\pm$0.014 &0.014$\pm$0.002 &0.11$\pm$0.02 &0.46$\pm$0.14 &0.09$\pm$0.01 &0.08$\pm$0.01 &0.11$\pm$0.04 &0.09$\pm$0.01 &0.02$\pm$0.01 &0.04$\pm$0.01 &0.03$\pm$0.01 \\
RSF &0.021$\pm$0.006 &0.088$\pm$ 0.046 &0.018$\pm$0.002 &0.12$\pm$0.02 &0.47$\pm$0.14 &0.11$\pm$0.01 &0.09$\pm$0.02 &0.13$\pm$0.05 &0.10$\pm$0.01 &0.03$\pm$0.01 &0.05$\pm$0.01 &0.03$\pm$0.01 \\
\bottomrule
\end{tabular}}
\end{sc}
\end{small}
\end{center}
\vskip -0.1in
\end{table*}

We present, here, the selection of representative results of our experiments. We focus on the results based on FSM performances. Indeed, a small FSM, by definition (\ref{evalscore}), guarantees a small MISE of the CATE and a small PEHE.

Figure \ref{fig:alpha} shows the FSM of SurvCaus in function on $\gamma_{\text{wd}}$. For small $\gamma_{\text{wd}}\ll 0.01$, we notice that the FSM is relatively large, and it decreases until it reaches the minimum for $\gamma_{\text{wd}}$ around $0.01$, then FSM starts to increase, and it explodes around $\gamma_{\text{wd}}\simeq 1$. This shows a high sensitivity of our estimates to $\gamma_{\text{wd}}$. Note that the magnitude of $\gamma_{\text{wd}}$ also depends on $d_{\text{WD}}^{\text{init}}$, which increases linearly with the number of features, as shown in Figure \ref{fig:wdnfeatures}. 

We also noticed that when we trained our model without the Wasserstein distance penalty (i.e., we set $\gamma_{wd}$ to 0), the performance remains similar to our model with a penalty when the initial Wasserstein distance $d_{\text{WD}}^{\text{init}}$ is already relatively small. Yet, a drastic increase of FSM is observed when we increase the $d_{\text{WD}}^{\text{init}}$. Moreover, the convergence speed is a breakneck of the SurvCaus model compared to the SurvCaus$_0$ and baseline methods.

Figures \ref{fig:LS} and \ref{fig:NLS} show the FSM of SurvCaus and baseline methods in function on $d_{\text{WD}}^{\text{init}}$ for linear and non-linear synthetic data. For a small distance $d_{\text{WD}}^{\text{init}}$, the baseline methods remain rather close in terms of FSM to ours.  Yet as soon as the $d_{\text{WD}}^{\text{init}}$ increases, we see in both linear and non-linear simulation schemes, a very strong increase of FSM for CoxPH and DeepSurv.  Surv-BART, RSF and EST remain relatively close in terms of FSM to our method which outperforms them all.

Tables \ref{table:linear} and \ref{table:nonlinear} show the MCATE, MPEHE and FSM of SurvCaus and baseline methods in the linear and non-linear schemes. 
We compare the means ($\pm$ standard deviations (sd)) of the MCATE, MPEHE, and FSM (lower the better) on the validation sets. We statistically compared the performances of SurvCaus over the five other methods using a bilateral Wilcoxon \cite{Wilcoxon1945IndividualCB} signed-rank test. In the results, $\dagger$ indicates the performance difference between SurvCaus and the method from the state-of-the-art is insignificant (i.e. p-value greater than 0.05). For simplicity of notation, significant results of p-value $< 0.05$ are not marked.

Our method outperforms baseline methods in both linear (see Table \ref{table:linear}) and nonlinear (see Table \ref{table:nonlinear}) simulation schemes, performances are ranked in the order: SurvCaus $\succeq $  Surv-BART $\succeq $ RSF $\succeq $  EST $\succeq $  CoxPH $\succeq $  DeepSurv.

SurvBART, RSF, and EST are relatively similar approaches that explain their similar performance. We also notice that CoxPH works well in the linear schema for small wd distances and vice versa.

Noting that the initial distances $d_{\text{WD}}^{\text{init}}$ corresponding to the data in the tables  \ref{table:linear} and \ref{table:nonlinear} are calculated after normalization (which largely decreases the distance) of the data, compared to the figures \ref{fig:LS} and \ref{fig:NLS} where they are calculated before normalization on different simulated data obtained with the same simulation scheme by increasing only the parameter $p_{wd}$. We note that baseline methods are sensitive to simulation parameters for the treated and untreated data sets, as the time horizon of the outcome for $\bar{F}_1$ is not always equal to that of $\bar{F}_0$. Our model outperforms the baseline methods because it considers the entire factual time horizon. So, we selected parameters that allow us to have two survival functions with the same horizon time for fair comparisons.

\section{Conclusion and discussion }
We present SurvCaus, a novel method to estimate individual treatment effects in a survival context setting. Our approach uses representation balancing and reweighing techniques to estimate survival functions and the CATE at the individual level by aligning factual and counterfactual distributions over a latent space.
We showed that the baseline methods are very deficient if they are trained on the whole dataset, with the treatment as a covariate. 

We first established theoretical guarantees for our algorithm, generalizing the work of~\cite{johansson2020generalization} to non-quadratic losses. In addition, we show that our algorithm significantly outperforms baseline methods on both synthetic and real datasets in both linear and nonlinear contexts. This is in adequacy with our theoretical findings of Section~\ref{sec:bounds}. 

The choice of discretization is essential, indeed we observed that the inverse discretization given by the Kaplan-Meier estimator with linear scheme interpolation performs better than the regular discretization, which validates the findings of \cite{kvamme2019continuous}. The performances are also sensitive to the $N_{\text{durations}}$ parameter, which motivates to consider in the future a penalty that automatically chooses the optimal number of subdivisions, $N_{\text{durations}}$ in the spirit of~\cite{cottin2021idnetwork}.  

We show that an increase in the distance between the distributions of the features in treated and untreated groups (in terms of Wasserstein distance) favors our method over baseline methods. We also show that the model performances are sensitive to $\gamma_{wd}$. We plan to investigate more the effect of $\gamma_{wd}$ from a theoretical perspective.

We plan to generalize our theoretical arguments to other settings, such as classification or situations with more than two lines of treatment, in future research.


\bibliographystyle{unsrt}  
\bibliography{main}

\begin{thebibliography}{10}

\bibitem{Brian2021IndividualComparisons}
Brian~G. Vegetabile.
\newblock On the distinction between "conditional average treatment effects"
  (cate) and "individual treatment effects" (ite) under ignorability
  assumptions, 2021.

\bibitem{Banerjee2009TheEconomics}
Abhijit~V Banerjee and Esther Duflo.
\newblock {The Experimental Approach to Development Economics}.
\newblock 2009.

\bibitem{SibekoExperimentalAfrica}
Goodman Sibeko and Dan~J Stein.
\newblock Experimental research: Randomised control trials to evaluate
  task-shifting interventions book title: Transforming research methods in the
  social sciences book subtitle: Case studies from south africa.

\bibitem{steinberg2004new}
Paul~F Steinberg.
\newblock {New approaches to causal analysis in policy research}.
\newblock 2004.

\bibitem{morrison2016searching}
Keith Morrison and Greetje van~der Werf.
\newblock {Searching for causality in educational research}.
\newblock {\em Taylor {\&} Francis}, 2016.

\bibitem{Tricoci2009ScientificGuidelines}
Pierluigi Tricoci, Joseph~M. Allen, Judith~M. Kramer, Robert~M. Califf, and
  Sidney~C. Smith.
\newblock {Scientific evidence underlying the ACC/AHA clinical practice
  guidelines}.
\newblock {\em JAMA - Journal of the American Medical Association},
  301(8):831--841, 2 2009.

\bibitem{Greenland2001ConfoundingIH}
Sander Greenland and Hal Morgenstern.
\newblock Confounding in health research.
\newblock {\em Annual review of public health}, 22:189--212, 2001.

\bibitem{Cox1972RegressionMA}
D.~R. Cox.
\newblock Regression models and life tables (with discussion.
\newblock 1972.

\bibitem{Wei1992TheAF}
L~J Wei.
\newblock The accelerated failure time model: a useful alternative to the cox
  regression model in survival analysis.
\newblock {\em Statistics in medicine}, 11 14-15:1871--9, 1992.

\bibitem{Ishwaran2008RandomSF}
Hemant Ishwaran, Udaya~B. Kogalur, Eugene~H. Blackstone, and Michael~S. Lauer.
\newblock Random survival forests.
\newblock {\em The Annals of Applied Statistics}, 2:841--860, 2008.

\bibitem{Sparapani2016NonparametricSA}
Rodney Sparapani, Brent~R Logan, Robert~E. McCulloch, and Purushottam~W. Laud.
\newblock Nonparametric survival analysis using bayesian additive regression
  trees (bart).
\newblock {\em Statistics in medicine}, 35 16:2741--53, 2016.

\bibitem{Cui2020EstimatingHT}
Yifan Cui, Michael~R. Kosorok, Stefan Wager, and Ruoqing Zhu.
\newblock Estimating heterogeneous treatment effects with right-censored data
  via causal survival forests.
\newblock {\em ArXiv}, abs/2001.09887, 2020.

\bibitem{Henderson2018IndividualizedTE}
Nicholas~C Henderson, Thomas~A. Louis, Gary~L. Rosner, and Ravi Varadhan.
\newblock Individualized treatment effects with censored data via fully
  nonparametric bayesian accelerated failure time models.
\newblock {\em Biostatistics}, 2018.

\bibitem{Hu2021EstimatingHS}
Liangyuan Hu, Jiayi Ji, and Fan Li.
\newblock Estimating heterogeneous survival treatment effect in observational
  data using machine learning.
\newblock {\em Statistics in medicine}, 2021.

\bibitem{Johansson2020GeneralizationEffects}
Fredrik~D. Johansson, Uri Shalit, Nathan Kallus, and David Sontag.
\newblock {Generalization bounds and representation learning for estimation of
  potential outcomes and causal effects}, 2020.

\bibitem{chapfuwa2020survival}
Paidamoyo Chapfuwa, Serge Assaad, Shuxi Zeng, Michael Pencina, Lawrence Carin,
  and Ricardo Henao.
\newblock Survival analysis meets counterfactual inference.
\newblock {\em arXiv preprint arXiv:2006.07756}, 2020.

\bibitem{Shalit2017EstimatingIT}
Uri Shalit, Fredrik~D. Johansson, and David~A. Sontag.
\newblock Estimating individual treatment effect: generalization bounds and
  algorithms.
\newblock In {\em ICML}, 2017.

\bibitem{Rubin2005CausalDecisions}
Donald~B. Rubin.
\newblock {Causal inference using potential outcomes: Design, modeling,
  decisions}.
\newblock {\em Journal of the American Statistical Association}, 100(469),
  2005.

\bibitem{ImbensRecentDI}
Guido Imbens and Jeffrey~M. Wooldridge.
\newblock Recent developments in the econometrics of program evaluation.
\newblock {\em IZA Institute of Labor Economics Discussion Paper Series}.

\bibitem{Pearl2000CausalityMR}
Judea Pearl.
\newblock Causality: Models, reasoning and inference.
\newblock 2000.

\bibitem{Cole2004AdjustedSC}
Stephen~R. Cole and Miguel~A. Hern{\'a}n.
\newblock Adjusted survival curves with inverse probability weights.
\newblock {\em Computer methods and programs in biomedicine}, 75 1:45--9, 2004.

\bibitem{Daz2019StatisticalIF}
Iv{\'a}n D{\'i}az.
\newblock Statistical inference for data-adaptive doubly robust estimators with
  survival outcomes.
\newblock {\em Statistics in medicine}, 38 15:2735--2748, 2019.

\bibitem{klein2003survival}
John~P Klein and Melvin~L Moeschberger.
\newblock {\em Survival analysis: techniques for censored and truncated data},
  volume 1230.
\newblock Springer, 2003.

\bibitem{Hill2011BayesianInference}
Jennifer~L. Hill.
\newblock {Bayesian nonparametric modeling for causal inference}.
\newblock {\em Journal of Computational and Graphical Statistics}, 20(1), 2011.

\bibitem{tsybakov2003introduction}
Alexandre~B Tsybakov.
\newblock {\em Introduction {\`a} l'estimation non param{\'e}trique},
  volume~41.
\newblock Springer Science \& Business Media, 2003.

\bibitem{Shimodaira2000ImprovingPI}
Hidetoshi Shimodaira.
\newblock Improving predictive inference under covariate shift by weighting the
  log-likelihood function.
\newblock {\em Journal of Statistical Planning and Inference}, 90:227--244,
  2000.

\bibitem{johansson2020generalization}
Fredrik~D Johansson, Uri Shalit, Nathan Kallus, and David Sontag.
\newblock Generalization bounds and representation learning for estimation of
  potential outcomes and causal effects.
\newblock {\em arXiv preprint arXiv:2001.07426}, 2020.

\bibitem{Mller1997IntegralPM}
Alfred M{\"u}ller.
\newblock Integral probability metrics and their generating classes of
  functions.
\newblock {\em Advances in Applied Probability}, 29:429--443, 1997.

\bibitem{Gretton2012ATest}
Arthur Gretton, Karsten~M. Borgwardt, Malte~J. Rasch, Bernhard Sch{\"{o}}lkopf,
  and Alexander Smola.
\newblock {A kernel two-sample test}, 2012.

\bibitem{Villani2007OptimalNew}
Cédric Villani.
\newblock {Optimal Transport Old and New}.
\newblock {\em Media}, 338, 2007.

\bibitem{Arjovsky2017WassersteinG}
Mart{\'i}n Arjovsky, Soumith Chintala, and L{\'e}on Bottou.
\newblock Wasserstein gan.
\newblock {\em ArXiv}, abs/1701.07875, 2017.

\bibitem{Pinetz2019OnTE}
Thomas Pinetz, Daniel Soukup, and Thomas Pock.
\newblock On the estimation of the wasserstein distance in generative models.
\newblock In {\em GCPR}, 2019.

\bibitem{vapnik1999nature}
Vladimir Vapnik.
\newblock {\em The nature of statistical learning theory}.
\newblock Springer science \& business media, 1999.

\bibitem{Cortes2010LearningWeighting}
Corinna Cortes, Yishay Mansour, and Mehryar Mohri.
\newblock {Learning bounds for importance weighting}.
\newblock In {\em Advances in Neural Information Processing Systems 23: 24th
  Annual Conference on Neural Information Processing Systems 2010, NIPS 2010},
  2010.

\bibitem{Sriperumbudur2009OnIP}
Bharath~K. Sriperumbudur, Kenji Fukumizu, Arthur Gretton, Bernhard Scholkopf,
  and Gert R.~G. Lanckriet.
\newblock On integral probability metrics, $\phi$-divergences and binary
  classification.
\newblock {\em arXiv: Information Theory}, 2009.

\bibitem{kvamme2019continuous}
Håvard Kvamme and Ørnulf Borgan.
\newblock Continuous and discrete-time survival prediction with neural
  networks, 2019.

\bibitem{lee2018deephit}
Changhee Lee, William~R Zame, Jinsung Yoon, and Mihaela van~der Schaar.
\newblock Deephit: A deep learning approach to survival analysis with competing
  risks.
\newblock In {\em Thirty-second AAAI conference on artificial intelligence},
  2018.

\bibitem{Cuturi2013SinkhornDL}
Marco Cuturi.
\newblock Sinkhorn distances: Lightspeed computation of optimal transportation
  distances.
\newblock {\em arXiv: Machine Learning}, 2013.

\bibitem{Buchanan2014WorthTW}
Ashley~L Buchanan, Michael~G. Hudgens, Stephen~R. Cole, Bryan Lau, and
  Adaora~A. Adimora.
\newblock Worth the weight: using inverse probability weighted cox models in
  aids research.
\newblock {\em AIDS research and human retroviruses}, 30 12:1170--7, 2014.

\bibitem{Katzman2018DeepSurvPT}
Jared Katzman, Uri Shaham, Alexander Cloninger, Jonathan Bates, Tingting Jiang,
  and Yuval Kluger.
\newblock Deepsurv: personalized treatment recommender system using a cox
  proportional hazards deep neural network.
\newblock {\em BMC Medical Research Methodology}, 18, 2018.

\bibitem{Geurts2006ExtremelyRT}
Pierre Geurts, Damien Ernst, and Louis Wehenkel.
\newblock Extremely randomized trees.
\newblock {\em Machine Learning}, 63:3--42, 2006.

\bibitem{Bergstra2012RandomSF}
James Bergstra and Yoshua Bengio.
\newblock Random search for hyper-parameter optimization.
\newblock {\em J. Mach. Learn. Res.}, 13:281--305, 2012.

\bibitem{Weinstein2013TheCG}
John~N. Weinstein, Eric~A. Collisson, Gordon~B. Mills, Kenna R.~Mills Shaw,
  Bradley~A Ozenberger, Kyle Ellrott, Ilya Shmulevich, Chris Sander, and
  Joshua~M. Stuart.
\newblock The cancer genome atlas pan-cancer analysis project.
\newblock {\em Nature Genetics}, 45:1113--1120, 2013.

\bibitem{1990SUPPORTST}
Support: Study to understand prognoses and preferences for outcomes and risks
  of treatments. study design.
\newblock {\em Journal of clinical epidemiology}, 43 Suppl:1S--123S, 1990.

\bibitem{Curtis2012TheGA}
Christina Curtis, Sohrab~P. Shah, Suet-Feung Chin, Gulisa Turashvili, Oscar~M.
  Rueda, Mark~J. Dunning, Doug Speed, Andy~G. Lynch, Shamith~A. Samarajiwa,
  Yinyin Yuan, Stefan Gr{\"a}f, Gavin Ha, Gholamreza Haffari, Ali Bashashati,
  Roslin Russell, Steven McKinney, Anita Langer{\o}d, Andrew~R. Green, Elena
  Provenzano, Gordon~C. Wishart, Sarah~E. Pinder, Peter~H. Watson, Florian
  Markowetz, Leigh Murphy, Ian~O. Ellis, Arnie Purushotham, Anne-Lise
  B{\o}rresen-Dale, James~D. Brenton, Simon Tavar{\'e}, Carlos Caldas, and
  Samuel Aparicio.
\newblock The genomic and transcriptomic architecture of 2,000 breast tumours
  reveals novel subgroups.
\newblock {\em Nature}, 486:346 -- 352, 2012.

\bibitem{Wilcoxon1945IndividualCB}
Frank. Wilcoxon.
\newblock Individual comparisons by ranking methods.
\newblock {\em Biometrics}, 1:196--202, 1945.

\bibitem{cottin2021idnetwork}
Aziliz Cottin, Nicolas P{\'e}cuchet, Marine Zulian, Agathe Guilloux, and
  Sandrine Katsahian.
\newblock Idnetwork: A deep illness-death network based on multi-state event
  history process for disease prognostication.
\newblock page to appear, 2022.

\bibitem{Shimodaira2000ImprovingFunction}
Hidetoshi Shimodaira.
\newblock {Improving predictive inference under covariate shift by weighting
  the log-likelihood function}.
\newblock {\em Journal of Statistical Planning and Inference}, 90(2), 2000.

\bibitem{ROSENBAUM}
Paul~R. Rosenbaum and Donald~B. Rubin.
\newblock {The central role of the propensity score in observational studies
  for causal effects}.
\newblock {\em Biometrika}, 70(1):41--55, 04 1983.

\end{thebibliography}

\newpage
\appendix
\onecolumn

\section{Details for Section~\ref{sec:problemform}}\label{app:sec3}
Under Assumption \ref{ass:ignor_pos} and \ref{ass:indep} the distribution of $Y^c(t), \delta(t)$ conditionally to $X=x$  on $[0,\tau _{min}] \times \{0,1\}$, is given by,
 
\begin{equation}
\Big[ f^{\star}_{t}( x,s)\overline{H}^{\star}_{t}( x,s)\Big]^{d }\Big[ h^{\star}_{t}( x,s)\overline{F}^{\star}_{t}( x,s)\Big]^{1-d }
\end{equation}

For a candidate $f_t(x,\cdot)$, the partial negative log-likelihood associated with the observation on $Y^c(t), \delta(t)$ is given by : 
\begin{equation}\label{eq:survloss}
    L\big(x, y^{c}(t), \delta(t), f_{t}\big)=-\delta(t) \log f_{t}\big(x, y^{c}(t)\big)  -(1-\delta(t)) \log \Big(\bar{F}_{t}\big(x, y^{c}(t)\big)\Big)
\end{equation}
see, e.g. \cite{klein2003survival} for details on the partial likelihood.

Our pointwise loss, under ignorability, is then given by: 
\begin{align*}
\ell_{f_{t}}(x)&= \mathbb{E}_{(Y^c,\delta) \mid X,T}\Big[L\Big( X,Y^{c} ,\delta ,f_t\Big) \mid X=x, T=t\Big]
\\
&=\mathbb{E}_{(Y^c(t),\delta(t)) \mid X}\Big[L\Big( X,Y^{c}(t)  ,\delta(t) ,f_t\Big) \mid X=x\Big]
\\
& = - \int _{0}^{\tau_{min}}  f^{\star}_{t}( x,s)\overline{H}^{\star}_{t}( x,s) \log f_t(x,s) ds  -  \int _{0}^{\tau_{min}} h^{\star}_{t}( x,s)\overline{F}^{\star}_{t}( x,s) \log \overline{F}_{t}( x,s)ds.
\end{align*}
As a consequence, the Kullback-Leibler divergence, that we defined in Equation~\eqref{eqn:kullback}, can be written as
\begin{align}
  \KL_x \Big( f^{\star}_{t} ||f_{t}\Big) &= \ell _{f_{t}}( x) -\ell _{f^{\star}_{t}}( x)  \\
  &= \int _{0}^{\tau_{min}}  f^{\star}_{t}( x,s)\overline{H}^{\star}_{t}( x,s) \log \frac{f_t^{\star}(x,s)}{f_t(x,s)} ds  +  \int _{0}^{\tau_{min}} h^{\star}_{t}( x,s)\overline{F}^{\star}_{t}( x,s) \log \frac{\overline{F}_{t}^{\star}( x,s)}{\overline{F}_{t}( x,s)}ds.
\end{align}

Now, returning to the CATE definition (Definition \ref{def:CATE}), we can write
\begin{align*}
  \Big|\text{CATE}( f_{0} ,f_{1} ,x) -\text{CATE}\Big( f^{\star}_{0} ,f^{\star}_{1} ,x\Big) \Big|&=\Big|\int_{0}^{\tau_{min}}\Big[ f_{0}( x,y) -f^{\star}_{0}( x,y)\Big] -\Big[ f_{1}( x,y) -f^{\star}_{1}( x,y)\Big] \Big|\\
&\leq \ \int_{0}^{\tau_{min}} |f_{0}( x,y) -f^{\star}_{0}( x,y) |\ \ +\ \ \int_{0}^{\tau_{min}} |f_{1}( x,y) -f^{\star}_{1}( x,y) |\ \\
&\leq \ 2\ \Big( d^{x}_{TV}\Big( f_{0} ,f^{\star}_{0}\Big) \ +\ d^{x}_{TV}\Big( f_{1} ,f^{\star}_{1}\Big)\Big) \ (\text{see Equation \ref{dtv}})
\end{align*}

Or 
\begin{equation}
\label{catedtv}
    \frac18\Big[\text{CATE}( f_{0} ,f_{1} ,x)  -\text{CATE}( f^{\star}_{0} ,f^{\star}_{1} , x) \Big]^2 \leq   \big(d^{x}_{TV}( f_{0} ,f^{\star}_{0})\big)^2  +  \big(d^{x}_{TV}\big( f_{1} ,f^{\star}_{1})\big)^2.
\end{equation}

We now need to bound the total-variation terms by means of Kullback-Leibler divergence. First, notice that we can write for $t=0,1$,

\begin{align*}
    \KL_x \Big( f^{\star}_{t} ||f_{t}\Big) = \sum_{d=0,1} \int _{0}^{\tau_{min}}  &\Big[ f^{\star}_{t}( x,s)\overline{H}^{\star}_{t}( x,s)\Big]^{d }\Big[ h^{\star}_{t}( x,s)\overline{F}^{\star}_{t}( x,s)\Big]^{1-d } \\ 
    &\log \frac{ \Big[ f_{t}( x,s)\overline{H}_{t}( x,s)\Big]^{d }\Big[ h_{t}( x,s)\overline{F}_{t}( x,s)\Big]^{1-d }}{\Big[ f^{\star}_{t}( x,s)\overline{H}^{\star}_{t}( x,s)\Big]^{d }\Big[ h^{\star}_{t}( x,s)\overline{F}^{\star}_{t}( x,s)\Big]^{1-d }} \ ds.
\end{align*}

which is a divergence between $ g^{\star}_{t}( x,s,d)= \Big[ f^{\star}_{t}( x,s)\overline{H}^{\star}_{t}( x,s)\Big]^{d }\Big[ h^{\star}_{t}( x,s)\overline{F}^{\star}_{t}( x,s)\Big]^{1-d }$ and  $g_{t}( x,s,d)= \Big[ f_{t}( x,s)\overline{H}_{t}( x,s)\Big]^{d }\Big[ h_{t}( x,s)\overline{F}_{t}( x,s)\Big]^{1-d }$, that we omit by denoting it $ \KL_x \Big( f^{\star}_{t} ||f_{t}\Big)$.

To this divergence, we can apply the First Pinsker's inequality  (see~\cite{tsybakov2003introduction}),

\begin{equation}
    d^{x}_{TV}\Big( g_t^\star , g_t\Big) \leq \sqrt{\frac{1}{2}  \KL_x \Big( f^{\star}_{t} ||f_{t}\Big) }
\end{equation}
with 
\begin{align*}
    d^{x}_{TV}\Big( g_t^\star , g_t\Big) & = \frac{1}{2} \sum _{d=0,1}\int_{0}^{\tau_{min}} |g_t^\star(x,s,d) -g_t(x,s,d)|ds\\
&= \frac{1}{2}\int_{0}^{\tau_{min}}|f^{\star}_{t}( x,s)\overline{H}^{\star}_{t}( x,s) -f_{t}( x,y)\overline{H}^{\star}_{t}( x,s) |ds\\ &+ \frac{1}{2} \int_{0}^{\tau_{min}}|h^{\star}_{t}( x,s)\overline{F}^{\star}_{t}( x,s) -h^{\star}_{t}( x,s)\overline{F}_{t}( x,s) | ds \\
&=\frac{1}{2} \int_{0}^{\tau_{min}}|f^{\star}_{t}( x,s) -f_{t}(x,s) |\ \overline{H}^{\star}_{t}( x,s)ds  +  \frac{1}{2} \int_{0}^{\tau_{min}}|\overline{F}^{\star}_{t}( x,s) -\overline{F}_{t}( x,s) |\ h^{\star}_{t}( x,s)ds\\
&\geq \frac{1}{2} \ \Big(\int_{0}^{\tau_{min}}|f^{\star}_{t}( x,s) -f_{t}(x,s) |ds  \Big)\overline{H}^{\star}_{t}( x,\tau _{min}) \ \ \text{as}\ \overline{H}^{\star}_{t}( x,\cdotp ) \ \searrow \\
&= d_{TV}^x \Big( f^{\star}_{t} ,f_{t} \Big) \cdot \overline{H}^{\star}_{t}( x,\tau _{min})  
\end{align*}

We just obtained that
\begin{equation}
    d_{TV}^x \Big( f^{\star}_{t} ,f_{t}\Big)  \leq \frac{1}{ \overline{H}^{\star}_{t}( x,\tau _{min})  } \ \sqrt{\frac{1}{2}  \KL_x \Big( f^{\star}_{t} ||f_{t}\Big) }.
\end{equation}
Together with equation \ref{catedtv} and integrating with respect to the distribution of $X$, this leads to
\begin{align*}
    \text{PEHE}(f_1,f_0) &=\E_{X\sim p^\star_{X}} \Big|\text{CATE}( f_{0} ,f_{1} ,X) -\text{CATE}\Big( f^{\star}_{0} ,f^{\star}_{1} , X\Big) \Big|^2 \\&\leq \ \frac{4}{\eta^2 } \Big(\E_{X\sim p^\star_{X}}\Big[\KL_X\Big( f^{\star}_{0} ||f_{0}\Big)\Big] +\E_{X\sim p^\star_{X}}\Big[\KL_X \Big( f^{\star}_{1} ||f_{1}\Big)\Big] \Big),
\end{align*}
where $\eta$ defined in Equation \ref{eta}. We give Theorem \ref{theo:ctrlpehe} with the definition of Equation \eqref{erdef} for $\ER(f_t)$.

\section{Details for Section 4}
\subsection{Importance-reweighing}\label{app:section4}

We proceed to show how the excess risk $\ER$ in hypothesis may be computed by re-weighting the factual excess risk $\ER_t$. This method is widely used in statistics and machine learning \cite{Shimodaira2000ImprovingFunction,Cortes2010LearningWeighting,ROSENBAUM}. Under assumption of overlap, for all $ t\in \{0,1\}$, $x \in \mathcal X$ and a weighting function $ w : \mathcal{X}  \to [ 0,1]$, we have: 
\begin{align}
    \ER^{w}_{t}( f_{t}) &=\int _{\mathcal{X}} w(x,t) \KL_{x}\Big( f^{\star}_t ||f_t\Big) p^\star_{X | T=t}(x)  dx\\
    &=\int _{\mathcal{X}} w(x,t)\frac{p^\star_{X | T=t}(x)}{p^\star_{X }(x)} \KL_{x}\Big( f^{\star}_t ||f_t\Big) p^\star_{X }(x) dx.
\end{align}
The equality $ \ER( f_{t})=  \ER^{w}_{t}( f_{t})$ holds if
\begin{equation}
w(x, t)=\frac{p^\star_{X}(x)}{p^\star_{X | T=t}(x) }= 
\frac{\mathbb P(T=t)}{(2t-1)(e^{\star}(x)-1) +1-t}
\end{equation}
by Bayes theorem, where $e^{\star}(x)=p^{\star}_{T | X=x}(t=1)$ is the true propensity score \cite{ROSENBAUM}.

Keeping the previous notations and denoting $\alpha _{t} =\mathbb P ( T=t)$, we have,
\begin{equation}\label{eqn:decom}
    \ER( f_{t}) =\alpha _{t} \ER_{t}( f_{t}) +( 1-\alpha _{t}) \ER_{1-t}( f_{t}).
\end{equation}

We notice that $O : w \mapsto \ER^{w}_{t}( f_{t})$ is a linear operator, with $O(1)=\ER_{t}( f_{t})$. We denote $\tilde{w}(x,t) =\alpha _{t} +( 1-\alpha _{t})  w(x,t)$, then, $O(\tilde{w})= \alpha_t O(1) + (1-\alpha_t) O(w) $. Therefore,
\begin{equation}\label{eqn:eqw}
    \ \ER^{\tilde{w}}_{t}( f_{t}) =\alpha _{t} \ER_{t}( f_{t}) +( 1-\alpha _{t}) \ER^{w}_{t}( f_{t}).
\end{equation}
From these two equations \ref{eqn:decom} and \ref{eqn:eqw}, we can easily obtain the desired result in lemma \ref{lemma:decom}.

\subsection{Balanced representation learning}\label{app:balancedproof}

The invertibility of $\phi$ guarantees the identifiability of the true $f_0^\star$, $f_1^\star$ and the CATE, i.e. the following assumptions are verified: $ \forall t\in \mathcal{T} : Y(t) \perp T \mid \phi(X) $ (Ignorability) and  $\forall z\in \mathcal{Z}: \mathbb{P}(T=t\mid \phi(X)=z)>0$ (Overlap) \cite{ImbensRecentDI,Pearl2000CausalityMR}.

We denote, for all $(z,t) \in \mathcal{Z} \times \mathcal{T}$,
\begin{align*}
    p^{\star,\phi}_{X | T=t}(z) &=p^\star_{X | T=t}(\psi(z))=p^\star_{\phi(X) | T=t}(z)\\
    p^{\star,\phi,w}_{X | T=t}(z) &=p^{\star,w}_{X | T=t}(\psi(z))=p^{\star,w}_{\phi(X) | T=t}(z).
\end{align*} 

\paragraph{Proof of Theorem \ref{thm:bound} }
We assume that $\exists A_\phi,B_\phi>0 : \forall z \in \mathcal{Z} : |J_{\psi }(z) |\leq A_{\phi }$ and $ z\mapsto\KL_{\psi(z)}\Big( f^{\star}_t ||f_t\Big)/B_{\phi} \in \mathcal{L}$, where $J_{\psi}$ is the Jacobean of the representation inverse $\psi$ and $\mathcal{L}$ is a reproducing kernel Hilbert space (RKHS) induced by a universal kernel \cite{Gretton2012ATest}.

We begin the proof by proofing the first inequation of Theorem \ref{thm:bound}. By the definition of $\Delta _{t}^w( f_t)$, we can write
\begin{align*} 
\Delta _{t}^w( f_t) & =  \ER_{1-t}( f_t) -\ER^{w}_{t}( f_t)\\
&= \int _{x\in \mathcal{X}} \KL_{x}\Big( f^{\star}_t ||f_t\Big)\Big[p^\star_{X | T=1-t}(x) -p^{\star,w}_{X | T=t}( x)\Big] dx\\
 & =\int _{z\in \mathcal{Z}} \KL_{\psi ( r)}\Big( f^{\star}_t ||f_t\Big)\Big[ p^{\star,\phi}_{X | T=1-t}(z) - p^{\star,\phi,w}_{X | T=t}(z)\Big] |J_{\psi }( r) |\ dz\\
 & \leq \ A_{\phi } \ \int _{z\in \mathcal{Z}} \KL_{\psi ( r)}\Big( f^{\star}_t ||f_t\Big)\Big[ p^{\star,\phi}_{X | T=1-t}(z) - p^{\star,\phi,w}_{X | T=t}(z)\Big]dz\\
 &\leq \underbrace{ A_{\phi } B_{\phi }}_{C_\phi} \ \ \int _{z\in \mathcal{Z}}\frac{\KL_{\psi ( r)}\Big( f^{\star}_t ||f_t\Big)}{B_{\phi}}\Big[ p^{\star,\phi}_{X | T=1-t}(z) - p^{\star,\phi,w}_{X | T=t}(z)\Big] dz\\
 & \leq C_{\phi } \ \sup _{g \in \mathcal{L}}\int _{z\in \mathcal{Z}} g(z) \Big[ p^{\star,\phi}_{X | T=1-t}(z) - p^{\star,\phi,w}_{X | T=t}(z)\Big] dz\\
 & \leq C_{\phi } \ \text{IPM}_{\mathcal{L}}\Big( p^{\star,\phi}_{X | T=1-t} ,p^{\star,\phi,w}_{X | T=t}\Big).
\end{align*}

Hence, with the decomposition obtained in lemma \ref{lemma:decom}, knowing that $1-\alpha_t = \alpha_{1-t}$, we have
\begin{equation}
\ER( f_t^\phi) \leq \ER^{\tilde{w}}_{t}( f_t^\phi) +\alpha _{1-t}C_{\phi } \ \text{IPM}_{\mathcal{L}}\Big(  p^{\star,\phi}_{X | T=1-t}, p^{\star,\phi,w}_{X | T=t}\Big).
\end{equation}
which gives the following bound for the PEHE
\begin{align*}
\label{motiveloss}
    \frac{\eta^2}{4} \cdot \text{PEHE}(f_1^\phi,f_0^\phi)  &\leq \ER(f_0^\phi) + \ER(f_1^\phi) \\
    & \leq \ER^{\tilde{w}_0}_{0}( f_0^\phi) +\ER^{\tilde{w}_1}_{1}( f_1^\phi) \\
    &+C_{\phi }  \Big[ \alpha _{1}\text{IPM}_{\mathcal{L}}\Big(  p^{\star,\phi}_{X | T=1}, p^{\star,\phi,w_0}_{X | T=0}\Big) + \alpha _{0}\text{IPM}_{\mathcal{L}}\Big(  p^{\star,\phi}_{X | T=0}, p^{\star,\phi,w_1}_{X | T=1}\Big)\Big].
\end{align*}
We have 
\begin{align*}
    \ER^{\tilde{w}}(f^\phi) &= \mathbb{E}_{X,T,Y^c,\delta \sim p^\star_{(X,T,Y^c,\delta)}}\Big(\tilde{w}(X,T)\Big[L(X,Y^c,\delta,f^\phi)-L(X,Y^c,\delta,f^{\star})\Big]\Big)\\
    &= R^{\tilde{w}}(f^\phi) - R^{\tilde{w}}(f^{\star}) \\
    &= \alpha_0 \ER^{\tilde{w}_0}_0( f_0^\phi) + \alpha_1 \ER^{\tilde{w}_1}_1( f_1^\phi).
\end{align*}
Next, given that $\alpha_0 + \alpha_1=1$, we obtain
\begin{equation*}
    \ER^{\tilde{w}_0}_0( f_0^\phi) + \ER^{\tilde{w}_1}_1( f_1^\phi) \leq \underbrace{\max(\frac{1}{\alpha_0},\frac{1}{\alpha_1})}_{=\beta >1}   \ER^{\tilde{w}}(f^\phi),
\end{equation*}
which gives,
\begin{align*}
        \frac{\eta^2}{4} \cdot \text{PEHE}(f_1,f_0)  &\leq \beta \Big( R^{\tilde{w}}(f^\phi) - R^{\tilde{w}}(f^{\star}) \Big)  \\
    &+C_{\phi }  \Big[ \alpha _{1}\text{IPM}_{\mathcal{L}}\Big(  p^{\star,\phi}_{X | T=1}, p^{\star,\phi,w_0}_{X | T=0}\Big) + \alpha _{0}\text{IPM}_{\mathcal{L}}\Big(  p^{\star,\phi}_{X | T=0}, p^{\star,\phi,w_1}_{X | T=1}\Big)\Big]\\
\end{align*}
Next, we bound the two IPM distances using the triangular inequality. Indeed, by adopting the notation $p_t = p^{\star,\phi}_{X | T=t}$ and $p_t^{w_t} = p^{\star,\phi,w_t}_{X | T=t}$  to simplify the proof, we have 
\begin{align*}
\text{IPM}_{\mathcal{L}}\big( p_{0} ,p_{1}^{w_{1}}\big) & \leq \text{IPM}_{\mathcal{L}}\big( p_{0} ,p_{0}^{w_{0}}\big) +\text{IPM}_{\mathcal{L}}\big( p_{0}^{w_{0}} ,p_{1}^{w_{1}}\big)\\
\text{IPM}_{\mathcal{L}}\big( p_{1} ,p_{0}^{w_{0}}\big) & \leq \text{IPM}_{\mathcal{L}}\big( p_{1} ,p_{1}^{w_{1}}\big) +\text{IPM}_{\mathcal{L}}\big( p_{1}^{w_{1}} ,p_{0}^{w_{0}}\big).
\end{align*}
Then, noting that $ \alpha _{0} +\alpha _{1} =1$,
\begin{equation*}
\alpha _{1} \text{IPM}_{\mathcal{L}}\big( p_{0} ,p_{1}^{w_{1}}\big) +\alpha _{0} \text{IPM}_{\mathcal{L}}\big( p_{1} ,p_{0}^{w_{1}}\big) \leq \text{IPM}_{\mathcal{L}}\big( p_{1}^{w_{1}} ,p_{0}^{w_{0}}\big) +\alpha _{1} \text{IPM}_{\mathcal{L}}\big( p_{0} ,p_{0}^{w_{0}}\big) +\alpha _{0} \text{IPM}_{\mathcal{L}}\big( p_{1} ,p_{1}^{w_{1}}\big).
\end{equation*}

Therefore, 
\begin{equation*}
\alpha _{1} \text{IPM}_{\mathcal{L}}\big( p_{0} ,p_{1}^{w_{1}}\big) +\alpha _{0} \text{IPM}_{\mathcal{L}}\big( p_{1} ,p_{0}^{w_{1}}\big) \leq \text{IPM}_{\mathcal{L}}\big( p_{1}^{w_{1}} ,p_{0}^{w_{0}}\big) + \text{IPM}_{\mathcal{L}}\big( p_{1} ,p_{1}^{w_{1}}\big) + \text{IPM}_{\mathcal{L}}\big( p_{0} ,p_{0}^{w_{0}}\big)
\end{equation*}
and finally, 
\begin{align*}
   \frac{\eta^2}{4\beta} \cdot \text{PEHE}(f_1,f_0)  \leq   \underbrace{R^{\tilde{w}}(f^\phi)   +  \frac{C_{\phi}}{\beta} \text{IPM}_{\mathcal{L}}\Big(  p^{\star,\phi,w_0}_{X | T=0},p^{\star,\phi,w_1}_{X | T=1}\Big)}_{\text{term of interest}} +\underbrace{\mathcal{D}_{\mathcal{L}}-R^{\tilde{w}}(f^{\star})}_{\text{ constant term in  $f^\phi$}}.    
\end{align*}

\subsection{Lemma from \cite{Sriperumbudur2009OnIP}}
We give in the following Lemma a result from \cite{Sriperumbudur2009OnIP} that allow us to bound the difference between the $\operatorname{IPM}_{\mathcal{L}}(p, q)$ and their equivalents taken at their empirical counterparts.
\begin{lemma}{\cite{Sriperumbudur2009OnIP}}
Let $\mathcal{X}$ be a measurable space. Suppose $k$ is a universal, measurable kernel such that $\sup _{x \in \mathcal{X}} k(x, x) \leq C \leq \infty$ and $\mathcal{L}$ the reproducing kernel Hilbert space induced by $k$, with $\nu=\sup _{x \in \mathcal{X}, f \in \mathcal{L}} f(x)<\infty$. Then, with $\hat{p}, \hat{q}$ the empirical counterparts distributions on $\mathcal{X}$ of $p$ and $q$, from $m$ and $n$ samples, and with probability at least $1-\delta$,
\begin{equation}
    \Big|\operatorname{IPM}_{\mathcal{L}}(p, q)-\operatorname{IPM}_{\mathcal{L}}(\hat{p}, \hat{q})\Big| \leq \sqrt{18 \nu^{2} \log \frac{4}{\delta} C}\Big(\frac{1}{\sqrt{m}}+\frac{1}{\sqrt{n}}\Big) .
\end{equation}
where,
\begin{equation*}
    \operatorname{IPM}_{\mathcal{L}}(\hat{p}, \hat{q})= \sup_{g \in \mathcal{L}}\Big |\frac{1}{m} \sum_{i=1}^m g(X_i^p) - \frac{1}{n} \sum_{i=1}^n g(X_i^q)  \Big |
\end{equation*}
when $\{X_i^p\}_{i=1}^m \overset{i.i.d}{\sim } p$ and $\{X_i^q\}_{i=1}^n \overset{i.i.d}{\sim } q$.
\end{lemma}

\section{Experiments}
\label{app:experiments}
\subsection{Prediction task and benchmark}

\paragraph{Individual CATE predictions}
We define the predictions of interest in this subsection based on the time scales defined below. Based on our network's output, we can define an empirical version of the CATE by
\begin{equation*}
   \widehat{\CATE}(x,y) =\sum ^{m+1}_{j=k(y)+ 1} \sigma ^{1}_{j}( \Psi ,\phi ,x) - \sigma ^{0}_{j}( \Psi ,\phi ,x).
\end{equation*}

\paragraph{Interpolation for Continuous-Time Predictions}
For a continuous time $y \in\Big(\tau_{j-1}, \tau_{j}\Big]$, the linear interpolation of the discrete survival function takes the shape
\begin{equation*}
    \bar{F}(y)=\bar{F}\Big(\tau_{j-1}\Big)+\Big[\bar{F}\Big(\tau_{j}\Big)-\bar{F}\Big(\tau_{j-1}\Big)\Big] \frac{y-\tau_{j-1}}{\Delta \tau_{j}}
\end{equation*}
where $\Delta \tau_{j}=\tau_{j}-\tau_{j-1}$. This implies that in this interval, the density function $f(y)$ is constant. However, we have, 
\begin{equation*}
 f(y)=-\bar{F}^{\prime}(y)=\frac{\bar{F}\Big(\tau_{j-1}\Big)-\bar{F}\Big(\tau_{j}\Big)}{\Delta \tau_{j}}   
\end{equation*}
So we can now rewrite the expression of the survival function as
\begin{align*}
\bar{F}(y)&=\frac{\bar{F}\Big(\tau_{j-1}\Big) \tau_{j}-\bar{F}\Big(\tau_{j}\Big) \tau_{j-1}}{\Delta \tau_{j}}-\frac{\bar{F}\Big(\tau_{j-1}\Big)-\bar{F}\Big(\tau_{j}\Big)}{\Delta \tau_{j}} y \\
&= \alpha_{j}-\beta_{j} y.
\end{align*}

We used the log-sum-exp trick to rewrite the loss for numerical stability reasons, inspired by the PyCox implementation (see \cite{kvamme2019continuous})
\subsection{Results }

\paragraph{Simulation settings}
We consider two different simulation scenarios: 
\begin{itemize}
    \item \textbf{LS}  : Linear scheme where  $s(x) = x\beta ^\top$ and $\beta =\{ (-1)^j \exp(j/10) | j \in \{1,\cdots,p\}    \}$
    \item \textbf{NLS} : Non-linear scheme where $s(x) = \frac{1}{p-1} \sum_{j=1}^{p-1} \sin (x_j \times x_{j+1})$
\end{itemize}

We encountered some problems when choosing the value of $\beta$ according to the simulation schemes: for the linear case if $p_{wd}$ increases, the distance $d_{\text{WD}}^{\text{init}}$ increases, then the term  the $s(X)=X\beta$ explodes, which obliges us to normalize the data,  which implies the decrease of $d_{\text{WD}}^{\text{init}}$. Another solution is to normalize $\beta$ by dividing it over its norm. Several tests of the choice of $\beta$ were carried out by choosing to keep only a few active covariates (5 covariates) and put the remaining ones at zero.

For the non-linear case, for the same reasons, we were obliged to normalize the dataset, which limits our control of $d_{\text{WD}}^{\text{init}}$ via $p_{wd}$. It should be noted that, in state of the art, the choice of $\beta$ is generally made simply by taking the contribution of only a few covariates, which does not suit our approach, as we would like to test the influence of the $d_{\text{WD}}^{\text{init}}$ distance on the prediction performance. 

The parameter $\epsilon$ allows separating the two survival functions because it is the contribution of treatment on $\bar{F}_t$ . We fix $\alpha= 2$ and $\lambda=1$ and we control the censorship rate by varying $\kappa_{cens}$. In Table \ref{tab:param}, we list a non-exhaustive list of  considered parameters. 

\begin{figure}[t]
  \centering
  \includegraphics[scale=0.26]{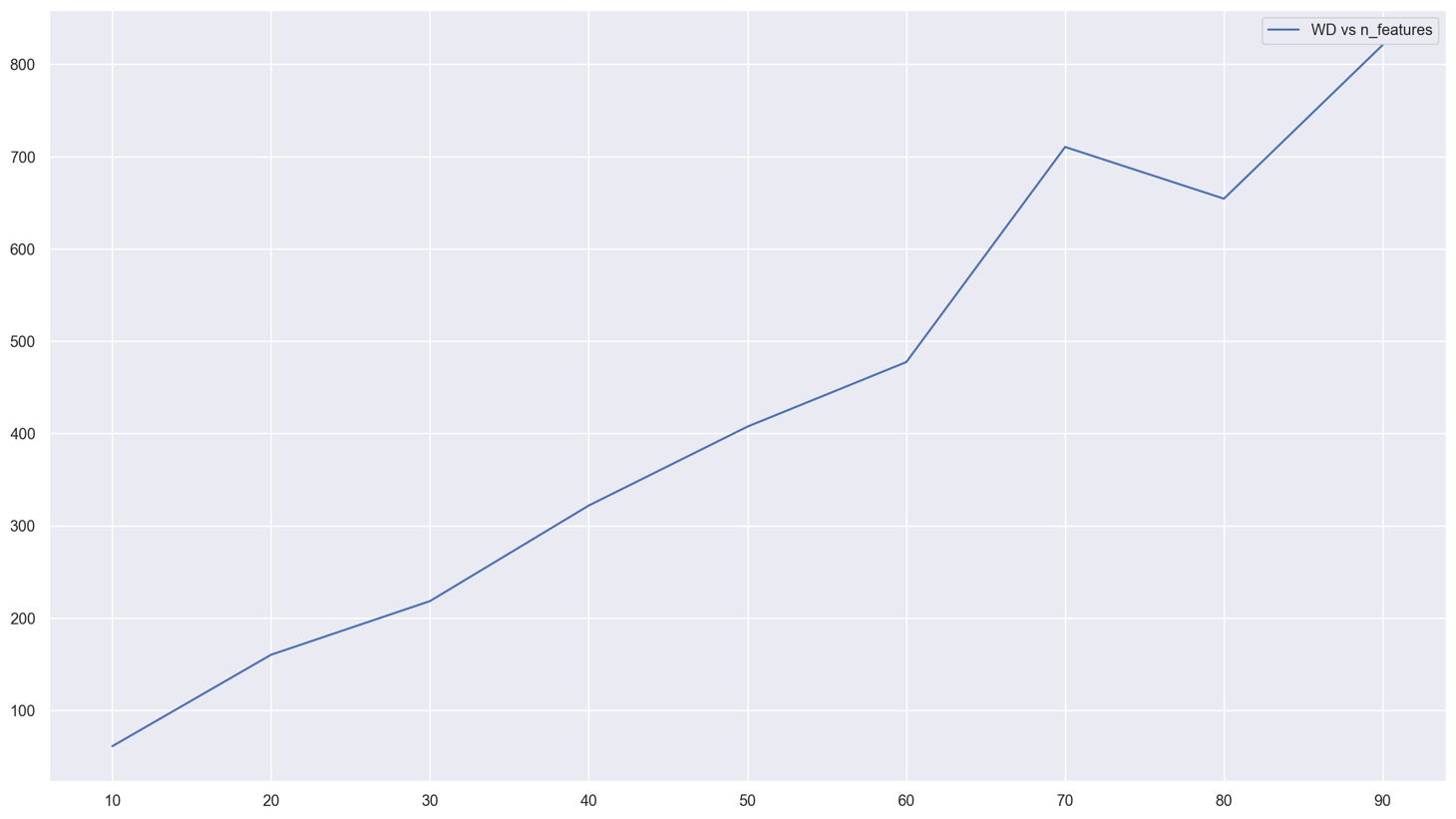}
    \caption{ $d_{\text{WD}}^{\text{init}}$ vs $N_{\text{features}}$}
  \label{fig:wdnfeatures}
\end{figure}

\begin{table}[h!]
  \centering
\begin{tabular}{lrrrrrrrrr}\toprule
\textbf{Scheme} &\textbf{N° samples} &\textbf{$N_{\text{features}}$} &\textbf{$p_{wd}$} &\textbf{$\rho$} &\textbf{$d_{\text{WD}}^{\text{init}}$} &\textbf{\% tt $=1$} &\textbf{\% event $=1$} &\textbf{$\epsilon$} \\\midrule
LS$_1$ &1000 &25 &4 &0.1 &36 &49 &73 &0.8 \\
LS$_2$ &1000 &25 &4 &0.1 &251 &49 &72 &0.8 \\
NLS&1000 &25 &10 &0.1 &662 &51 &50 &1.8 \\
\bottomrule
\end{tabular}
\caption{Some simulation parameters}
\label{tab:param}
\end{table}

\end{document}